\documentclass[lettersize,journal]{IEEEtran}
\usepackage{amsmath,amsfonts}
\usepackage{algorithmic}
\usepackage{algorithm}
\usepackage{array}
\usepackage{fontawesome5}
\usepackage[colorlinks=true, linkcolor=red, citecolor=ngreen, urlcolor=blue]{hyperref}
\usepackage[caption=false,font=normalsize,labelfont=sf,textfont=sf]{subfig}
\usepackage{textcomp}
\usepackage{multirow} 
\usepackage{stfloats}
\usepackage{url}
\usepackage{flushend}
\usepackage{verbatim}
\usepackage{graphicx}
\usepackage{relsize}
\usepackage{pgfplots}
\usepackage{colortbl}
\usepackage{cite}
\usepackage{float}
\usepackage{dsfont}
\usepackage{orcidlink}
\usepackage{arydshln}
\usepackage{enumitem}
\usepackage{float}
\usepackage{hyperref}
\usepackage{multirow}
\usepackage{caption}
\usepackage{amssymb}
\usepackage{booktabs}
\usepackage{bbm}
\usepackage{tikz}
\usepackage{bm}
\usepackage[normalem]{ulem}
\usepackage{xcolor}

\pgfplotsset{compat=1.18}

\begin{document}
\definecolor{ngreen}{RGB}{17, 173, 30}
\newcommand{\reddashedline}{\textcolor{red}{\rule[0.5ex]{0.3em}{1pt}\hspace{0.2em}\rule[0.5ex]{0.3em}{1pt}\hspace{0.2em}\rule[0.5ex]{0.3em}{1pt}\hspace{0.2em}\rule[0.5ex]{0.3em}{1pt}}}
\definecolor{mycolor_green}{RGB}{88, 142, 49}
\definecolor{mycolor_red}{RGB}{192, 0, 0}
\definecolor{mycolor_orange}{RGB}{242, 186, 2}
\newcommand{\cyh}[1]{{\color{red}{[YH: #1]}}}
\newcommand{\tychen}[1]{{\color{orange}{[tychen: #1]}}}
\newcommand{\hbliu}[1]{{\color{blue}{[hbliu: #1]}}}
\newcommand{\mycite}[1]{\textcolor{red}{#1}}
\setlength{\arrayrulewidth}{1mm}
\title{\textit{MECD+}: Unlocking Event-Level Causal Graph Discovery for Video Reasoning}

\author{Tieyuan Chen$^{\orcidlink{0009-0005-7939-7139}}$, Huabin Liu, Yi Wang, Yihang Chen, Tianyao He,\\ Chaofan Gan, Huanyu He, Weiyao Lin,~\IEEEmembership{Senior Member, IEEE}
\thanks{The paper is supported in part by the National Natural Science Foundation of China (No. 62325109, U21B2013), the Shanghai `The Belt and Road' Young Scholar Exchange Grant (24510742000), the National Key R\&D Program of China (Grant No. 2022ZD0160102), and the Zhongguancun Academy Project No.20240313.
\textit{(Corresponding authors: Huabin Liu, Weiyao Lin.)}}
\thanks{Tieyuan Chen, Huabin Liu, Weiyao Lin, Yihang Chen, Tianyao He, Chaofan Gan, and Huanyu He are with Shanghai Jiao Tong University, Shanghai, China. Yihang Chen is also with Monash University, Melbourne, Australia. Tieyuan Chen and Weiyao Lin are also with the Zhongguancun Academy, Beijing, China. E-mail:\{tieyuanchen, wylin\}@sjtu.edu.cn.}
\thanks{Yi Wang is with the Shanghai AI  Laboratory, Shanghai, China. E-mail: wangyi@pjlab.org.cn.}
}
\markboth{IEEE TRANSACTIONS ON PATTERN ANALYSIS AND MACHINE INTELLIGENCE}%
{Chen \MakeLowercase{\textit{et al.}}: MECD}

\maketitle

\begin{abstract}
Video causal reasoning aims to achieve a high-level understanding of videos from a causal perspective. 
However, it exhibits limitations in its scope, primarily executed in a question-answering paradigm and focusing on brief video segments containing isolated events and basic causal relations, lacking comprehensive and structured causality analysis for videos with multiple interconnected events. 
To fill this gap, we introduce a new task and dataset, \textbf{M}ulti-\textbf{E}vent \textbf{C}ausal \textbf{D}iscovery (MECD). It aims to uncover the causal relations between events distributed chronologically across long videos. 
Given visual segments and textual descriptions of events, MECD identifies the causal associations between these events to derive a comprehensive and structured event-level video causal graph explaining why and how the result event occurred. 
To address the challenges of MECD, we devise a novel framework inspired by the Granger Causality method, incorporating an efficient mask-based event prediction model to perform an \textit{Event Granger Test}. It estimates causality by comparing the predicted result event when premise events are masked versus unmasked. 
Furthermore, we integrate causal inference techniques such as front-door adjustment and counterfactual inference to mitigate challenges in MECD like causality confounding and illusory causality. Additionally, context chain reasoning is introduced to conduct more robust and generalized reasoning.
Experiments validate the effectiveness of our framework in reasoning complete causal relations, outperforming GPT-4o and VideoChat2 by 5.77\% and 2.70\%, respectively.
Further experiments demonstrate that causal relation graphs can also contribute to downstream video understanding tasks such as video question answering and video event prediction.
\end{abstract}


\section{Introduction}

\IEEEPARstart{V}{\lowercase{ideo}} causal reasoning aims to understand and analyze video content from a causal perspective. 
It requires models to comprehend temporal relationships, anticipate actions, and adapt to dynamic visual elements across frames. This capability is essential in various real-world applications, including autonomous driving~\cite{driving}, activity recognition~\cite{act}, video surveillance~\cite{automatic}, and even complex decision-making scenarios like robotic navigation~\cite{robot}. 
Among these applications, Video Question Answering (VQA)~\cite{SeViLA, NEXT, CLEVRER} represents one of the most prominent tasks in video causal reasoning, where models are tested on their causal ability to understand videos through causal questions such as explanations, predictions, and counterfactual assumptions. Traditional VQA tasks can be viewed as attempts to discover single causal links within videos, as they typically identify one reason from given multiple options that explain the event described in the question.

Recent advancements in VQA have expanded beyond traditional tasks by incorporating more sophisticated causal reasoning methods. 
Notable trends include enhancing causal chain inference from simple one-to-one relations to multi-to-one chains, as seen in multiple correct answers VQA~\cite{nmulti1, nmulti2, nmulti3}, and improving answer causal discovery through temporal grounding techniques, as seen in Grounded Video Question Answering (grounded VQA)~\cite{nextgqa, videotree, timecraft}.
Recent research also endeavors to infer a more comprehensive causal graph focusing on a particular event within its contexts~\cite{BiGED,VAR}. 


\begin{table*}[t]
    \centering
        \caption {\textbf{Video Reasoning Tasks Comparison.} Our MECD task uniquely enables the reasoning of a comprehensive causal graph at the event level through causal reasoning.}
    \resizebox{0.8\textwidth}{!}{
      \setlength{\tabcolsep}{0.9mm}
  \begin{tabular}{ccccccc}
    \toprule
    $\textbf{Task}$ & $\textbf{Work}$ & $\textbf{Causal Reasoning}$ & $\textbf{Event-Level}$ & $\textbf{Multi-Chains}$  & $\textbf{Complete Graph}$\\
    \midrule
    $\textbf{Common VQA}$ & Next-QA~\cite{NEXT} &\checkmark & &  & \\
    $\textbf{Multi-Answer VQA}$ & ProViQ~\cite{nmulti1} & & &\checkmark & \\
    $\textbf{Grounded VQA}$ &NextGQA~\cite{nextgqa}    &\checkmark &\checkmark & & \\
    $\textbf{Causal Event Prediction}$ &VAR~\cite{VAR}  & &\checkmark &\checkmark  &  \\
    $\textbf{Causal Relation Based VQA}$ &REXTIME~\cite{rextime}  &\checkmark &\checkmark &\checkmark  &  \\
    $\cellcolor{green!10}\textbf{Causal Graph Reasoning}$ &\cellcolor{green!10}\textbf{Our MECD}  &\cellcolor{green!10}\checkmark &\cellcolor{green!10} \checkmark &\cellcolor{green!10} \checkmark &\cellcolor{green!10}\checkmark  \\
    \bottomrule
  \end{tabular}
  \label{tab:task_diff}
}
\end{table*}

Despite recent advancements, as shown in Tab.~\ref{tab:task_diff}, current video causal reasoning tasks remain limited in scope, primarily focusing on QA-based approaches that discover a single causal relation within a video. 
However, when the scene to be analyzed is more complex, these tasks often fail to meet the required understanding standards. 
For example, in a video where a person slips and falls while carrying a tray of drinks, the cause might be combined with a wet floor, slippery shoes, and losing balance due to walking too quickly. 
Besides, these tasks often lack the ability to perform fine-grained event-level reasoning, which is crucial in real-world scenarios. Such detailed reasoning typically leads to clearer insights into video content. 
Most importantly, they fail to provide a comprehensive, structured causal representation. 
For instance, in traffic surveillance videos, a detailed analysis of events occurring at different times is essential to determine which events, or combinations of events, resulted in a specific accident among the chain of traffic accidents.

To address this gap, we set up a new task: \textbf{M}ulti-\textbf{E}vent \textbf{C}ausal \textbf{D}iscovery (MECD), which aims to uncover causal relations among events that distribute chronologically.

\begin{figure*}[t]
\begin{center}
\includegraphics[width=0.92\textwidth]{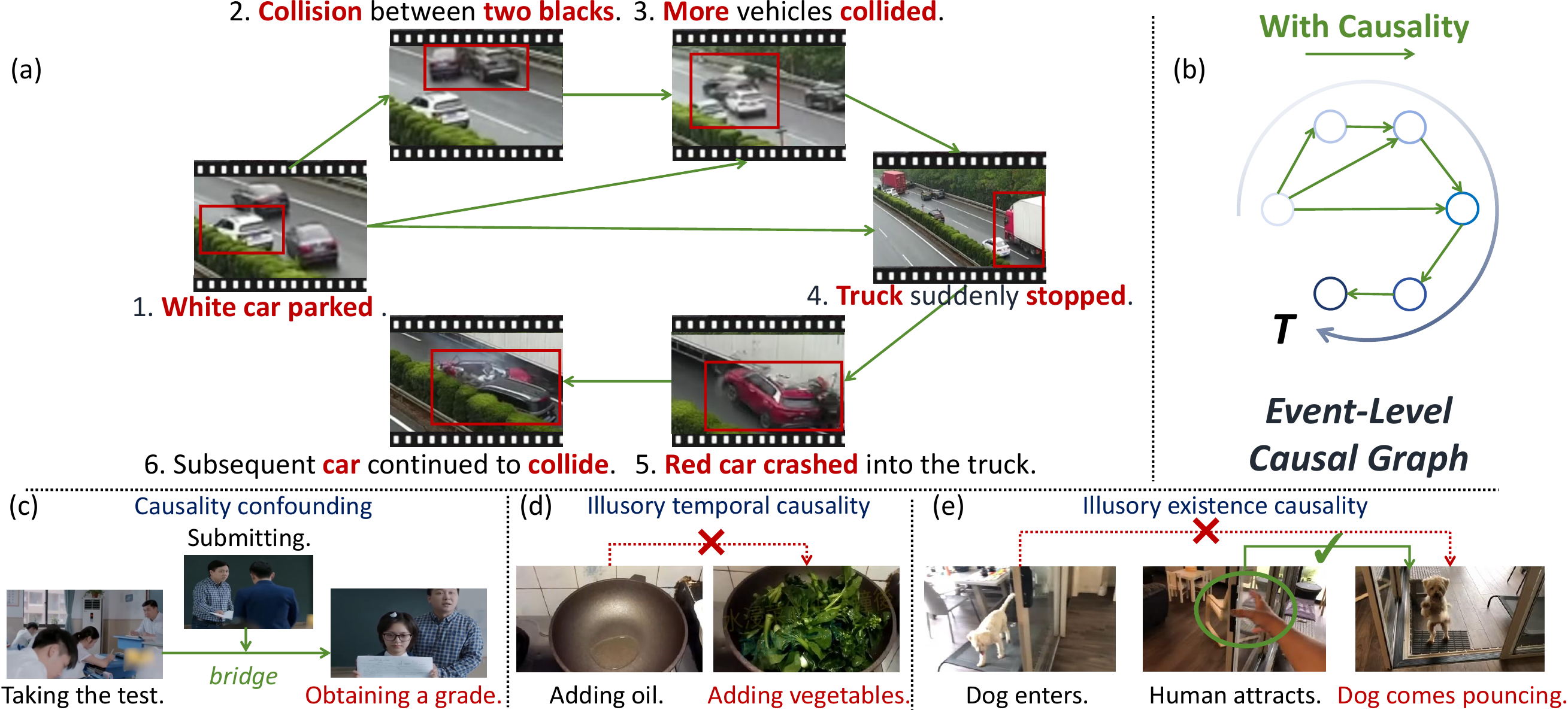}
\end{center}
\vspace{-5pt}
\caption{ {\textbf{(a): Illustration of Multi-Event Causal Discovery Task}}, where a complete causal graph of traffic surveillance videos is shown. Our task aims to determine whether a causal relation exists between events and outputs a structured causal graph in (b). (c): Example of causality confounding. 
 (d)\&(e): Illustration of illusory causality.}
\label{intro_figure}
\vspace{-5pt}
\end{figure*}

As illustrated in Fig.~\ref{intro_figure}, given multiple chronologically arranged event segments in a video along with their corresponding textual descriptions (Fig.~\ref{intro_figure} (a)), MECD requires identifying causal relations between these events to derive a comprehensive and structured event-level causal graph (Fig.~\ref{intro_figure} (b)), indicating why and how every event happens. 
Meanwhile, we contribute a new dataset for the training and evaluation of MECD by collecting moderately long-form videos involving multiple events and manually annotating the ground truth causal relation between any events pair. 

However, to our knowledge, no available solutions can directly comprehend causal relations at the event level, necessitating the development of a new framework.

To this end, we draw inspiration from the \textit{Granger Causality Method}~\cite{granger,granger2, granger3} for solution, which is widely used in traditional causal discovery for low-dimensional time-series data (e.g., stock prices, weather patterns). 
The main idea is that temporal causality can be effectively estimated by predictive ability. Specifically, applied to videos, if Event A occurs prior to Event B, we consider A to be a cause of B only if A could facilitate the prediction of B. 
We term this criterion the \textit{Event Causality Test}. However, compared to simple low-dimensional data, the inputs of MECD involve much more complex modalities, including both visual and textual content, which may introduce bias in the estimation of causality using such a predictive paradigm. 
Specifically, we observe that directly applying \textit{Event Causality Test} to video causal discovery presents two main problems:

(1) \textbf{Causality confounding} indicates that the original causal relations between events are disrupted or interfered with by other relay or adjacent events. Such confounding stems from the fact that many causal relations flow through an intermediary event that acts as a bridge.  
{For example, in Fig.~\ref{intro_figure}(c), “submitting the paper” merely mediates the effect of “taking the test” on “obtaining a grade”, yet may be mistakenly treated as the sole cause, thereby obscuring the true causal dependency.}

(2) \textbf{Illusory Causality}, including both illusory temporal and existence causality.  \texttt{Illusory temporal causality} exists when events exhibit a close correlation in temporal distribution.   Such correlation may mislead the test of real causality.     
{For example, in Fig.~\ref{intro_figure}(d), “adding oil when cooking” often precedes “adding vegetables to stir-fry”, yet the former does not cause the latter.}
As for \textit{illusory existence causality}, it occurs when some objects in early events may serve as necessary existence conditions of a later event. 
{For instance, as shown in Fig.~\ref{intro_figure}(e), while “a large brown dog enters the room” is a prerequisite for “the dog runs towards the camera”, the former does not cause the latter.}

Building upon the preceding discussion, we introduce a novel framework named VGCM (Video Granger Causality Model) to tackle MECD. 
This framework executes the \textit{Event Granger Test} via an efficient mask-based event prediction model. 
According to the Granger Causality theorem in machine learning~\cite{granger,  granger2, granger3}, if the inclusion of premise events can improve the prediction of a subsequent event, then they are causally related. 
{Besides, the video data used in this article is theoretically and rigorously compliant with the conditions for Granger Causality, as it represents a continuous-in-time, deterministic, and complete system where potential confounders are mitigated through causal inference.}
Following this theory, VGCM deduces the causality of a premise event by comparing the predicted features of the result event when the premise is either masked or unmasked. 
Furthermore, to mitigate the challenges of causality confounding and illusory causality discussed earlier, we integrate two additional causal inference techniques—front-door adjustment~\cite{causalinference, hanwangzhang} and counterfactual inference~\cite{causalinference,count2}—into our framework. 
Specifically, these techniques compensate for or remove the causal effects of previous or subsequent adjacent bridge events to eliminate confounding. 
Simultaneously, they address the illusory causality issue by incorporating the chain of thoughts~\cite{cot1, cot2, cotsurvey} and existence-only descriptions.

Furthermore, we improve model robustness by applying context chain reasoning in the \textit{Event Causality Test}, enhancing interactions among in-context causal chains. 
In reality, there are numerous instances where partial causes, when combined, contribute to the formation of the complete cause leading to the result.
To enable the model to learn this reasoning pattern commonly found in reality, we randomly mask multiple premise events during training.


Extensive experiments validate the effectiveness and generalizability of our proposed framework VGCM in discovering structured causal relations for given long-form videos. 
Simultaneously, it has been proven that the complete causal graph inferred by our VGCM significantly promotes the downstream video understanding tasks such as video question answering and video event prediction.

This manuscript extends our NeurIPS 2024 paper~\cite{MECD} and makes the following contributions: The release and deeper analysis of MECD+, a larger multi-video source dataset. An enhancement of the \textit{Event Causality Test} through robust context chain reasoning and efficient non-regressive complete graph reasoning. Additional metrics and further analysis of the MECD task, such as input modalities, hallucination issues, and downstream applications.

\section{Related work}
\noindent\textbf{Video Causal Reasoning.} Many existing tasks have tried to carry out causal reasoning in videos.
Among these, the most common task is Video Question Answering (VQA), aiming to deduce a reason according to the question, grounded-QA methods such as SeViLA, NextGQA and Momentor~\cite{SeViLA, nextgqa, momentor} grounded a single reason happens before the result. 
However, the traditional VQA task does not extend to abducting multiple reasons, merely abducting a single causal chain from reason to the provided result. 
{Some studies have enhanced video reasoning paths by introducing Chain-of-Thought (CoT) to better accomplish VideoQA~\cite{videocot, videoofthought, VideoEspresso} or Rumor Detection~\cite{Following}, however, these works remain constrained by single-chain causal inference.}
Recently, some methods, such as Glance~\cite{glance}, VideoTree~\cite{videotree}, and TimeCraft~\cite{timecraft} have constructed more detailed causal chains through reasoning from coarse-grained to fine-grained levels or bidirectional paths.
Although these works improve the causal reasoning process, they only abduct the causal chain for a single result.

Beyond the above classical setting, multiple correct answers VQA~\cite{nmulti1, nmulti2, nmulti3} focuses on a more flexible setting which could exist multiple correct reasons corresponding to the result, however, most of the choices have correlations or paradoxes between them. Therefore this task still does not suggest a generalized complete causal graph.

Furthermore, many tasks are based on VQA for further causal reasoning attempts. 
Physical state reasoning task CLEVRER~\cite{CLEVRER} and CATER~\cite{CATER} explored causal reasoning based on physics and other basic laws in virtual scenes. However, they haven't been committed to extending to the general video. 
Neural-symbolic paradigm AAR~\cite{AAR} and LMLN~\cite{LMLN} symbolized data and derived inference rules using external knowledge. However, they can only reason within a defined symbol domain.
The most similar task VAR~\cite{VAR} predicted explanation events with premise events, and the causality is introduced during its prediction process. However, it hasn't been committed to discovering the complete causal graph.
Another concurrent study REXTIME~\cite{rextime} determined the causal relation between two events in a video and then designed QA questions targeted at the relations. However, it only reasoned through incomplete causal chains.  {Moreover, unlike the causal relation reasoning tasks mentioned above, the video spatial relationship reasoning task~\cite{s-relation} aims at inferring the semantic connection between two objects formed by a specific action or state, rather than the causal relation that is our primary focus.}

Most of these tasks introduced above are coarse video-level reasoning tasks, ours is more fine-grained event-level reasoning. 
Besides, we devote ourselves to abducting a complete causal graph rather than causal chains related to a single event. 
In conclusion, to the best of our knowledge, current video reasoning tasks haven't been committed to discovering complete causality among complex multi-event videos. Consequently, there exists a necessity need for a more comprehensive task.

\noindent\textbf{Causal Discovery.} 
Current causal discovery predominantly focuses on two main application areas: time series statistical data and natural language processing (NLP).

The primary objective of causal discovery in the context of time series statistical data is to infer the causal relation between the independent variable and the dependent variable. 
Traditional causal discovery methods are mainly divided into three categories: Constraint-based, Score-based, and Granger Causality method. 
Constraint-based methods utilize conditional independence tests to identify causal relations~\cite{LPCMCI, PcGCE}, and Score-based methods search through the space of all possible causal structures to optimize a speciﬁed metric~\cite{NTS, DYNOTEARS}. The constraint-based and score-based methods require stringent assumptions about data distribution, which would severely limit their generalization capabilities. 
Besides the two methods above, the more widely used Granger Causality method discovers causal relations by calculating the degree to which the earlier occurred event contributes to predicting the latter occurred event~\cite{THP, GC-nsHP}. 

The task of causal reasoning in natural language processing (NLP) generally involves inferring the causal relations between words within a text paragraph. 
{TCR~\cite{TCR} proposes employing constrained conditional models and formulates the extraction of causal relations between events as an integer linear programming problem.}
The Following methods usually incorporate external knowledge, either through knowledge graphs or prompts.
For example, RFBFN~\cite{RFBFN} and CauSeRL~\cite{CauSeRL} utilized knowledge related to candidate causal events and external causal statements from the knowledge graph. 
However, these approaches are generally limited to handling sentences with a single causal relation pair. {GESI~\cite{GESI} further advances this by constructing heterogeneous graph-based models to capture more complex causal relations, potentially spanning across multiple paragraphs. However, GESI suffers from slower inference speeds and increased complexity, which can be prohibitive for higher-dimensional video data applications.}
{Additionally, following the Granger Causality method, CAUSE~\cite{CAUSE} proposes capturing event dependencies by fitting neural point processes and then employing axiomatic attribution to quantify the contribution of preceding events to subsequent predictions.}

Therefore, for the task of causal discovery in video, it is not feasible to simply apply methods toward time series statistical data or NLP directly. 
Instead, we leverage the principles of causal discovery and design a pipeline that satisfies the specific characteristics of video data. 
Specifically, due to the straightforward nature of Granger Causality and its potential for integration with research related to masks in video understanding, it is selected as the foundational principle for reasoning in video data.

\begin{figure}[t]
\begin{center}
\includegraphics[width=0.45\textwidth]{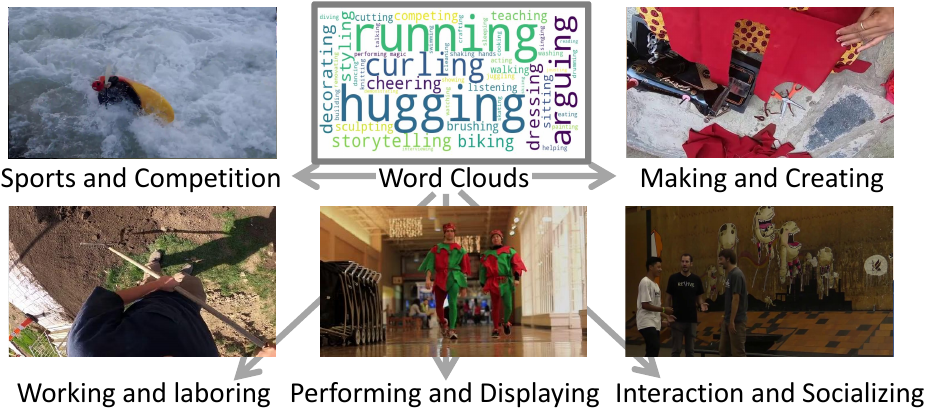}
\end{center}
\vspace{-5pt}
\caption{\textbf{Constitute of the MECD Dataset.} We present 5 main video categories and the verb. word cloud of the dataset.}
\label{fig: categories}
\vspace{-10pt}
\end{figure}

\section{Multi-Event Causal Discovery}
\begin{figure*}[t]
\begin{center}
\includegraphics[width=0.98\textwidth]{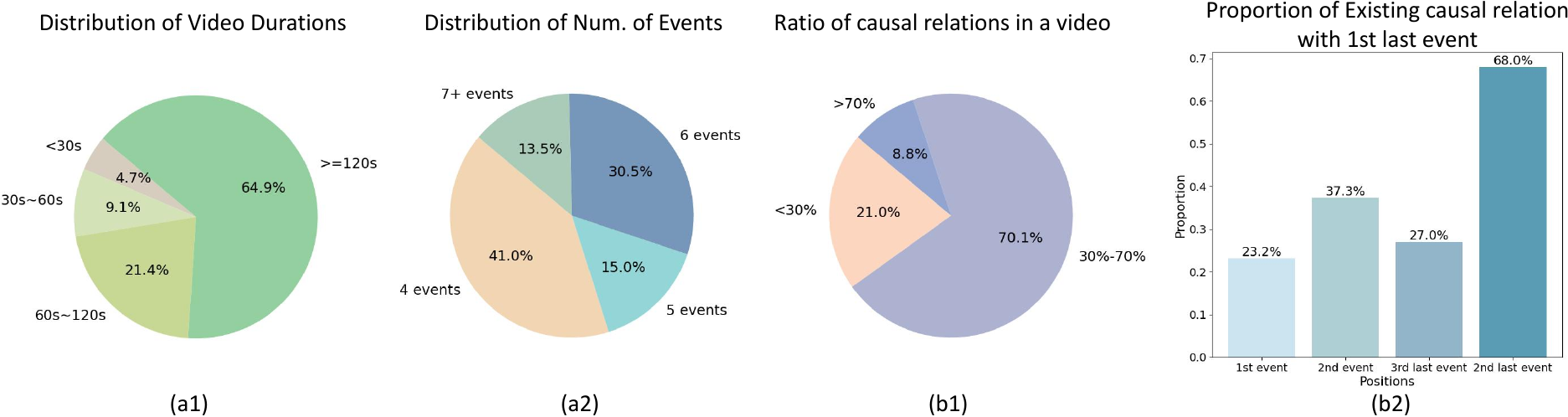}
\end{center}
\vspace{-10pt}
\caption{\textbf{Statistics of the MECD Dataset.} As shown in Figures (a1) and (a2), our dataset mainly analyzes videos that are longer than 120 seconds and contain five or more events. As shown in Figures (b1) and (b2), {the proportion of causal and non-causal relations between events in the video of MECD is relatively balanced}, moreover emphasizing the existence of causal relations between adjacent events.}
\label{fig: statistics}
\end{figure*}

To quantify the ability of causal discovery of a given model in multi-event videos, we propose the task of Multi-Event Causal Discovery (MECD). In formulation, given a video $\mathcal{E}$ that contains chronologically organized $N$ events, $\mathbb{E}:=\{e_1, \dots,  e_N\}$, this task aims at determining whether any previous event $e_n$ ($n<N$) has a causal relation with the last one (\textit{i.e.}, $e_N$). Specifically, an event $e_n=\{v_n,  c_n\}$ consists of a video clip $v_n$ and the corresponding caption $c_n$.
Without loss of generality, relations of previous events to the last one can be expressed as ${\boldsymbol r}=[r_1, \dots, r_{N-1}]$, where $r_k$ ($k<N$) is set to ``1'' to indicate the existence of $e_k$'s causal relation with $e_N$, and ``0'' otherwise.
Notably, this setting {\textbf{(defined as causal chains discovery)}} is generalizable to causal relations of any two events as long as we cut off the video and treat the latter one as the last event.

\subsection{Task Data}

\begin{figure*}[t]
\begin{center}
\includegraphics[width=0.8\textwidth]{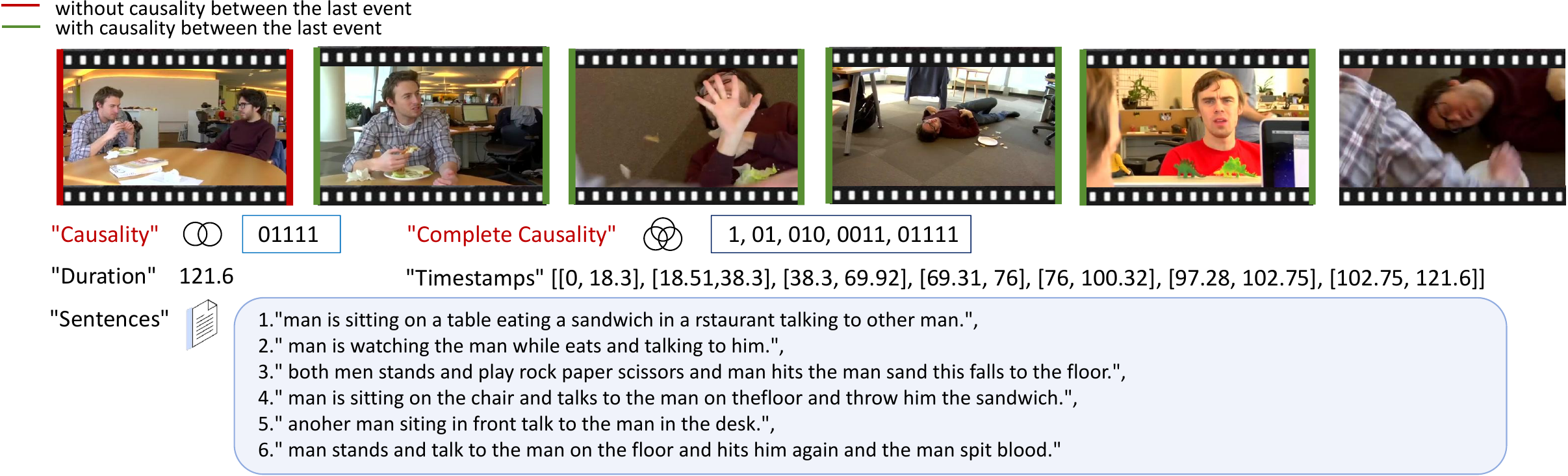}
\end{center}
\vspace{-10pt}
\caption{\textbf{Annotation Examples of the MECD Dataset.} Newly annotated attributes ``causality'', ``complete causality'' and the formerly existing attributes ``sentences'', ``duration'', and ``timestamps'' are shown along with the video frames.}
\label{fig:sup_annotations2}
\end{figure*}

\noindent\textbf{Data Source.}
The Multi Events Causal Discovery (MECD) task contains videos with multiple events and intricate causal relations. 
We carefully reorganize many widely used long-term daily videos including the ActivityNet Captions~\cite{ActivityNet}, EgoSchema~\cite{egoschema}, and NExTVideo~\cite{NEXT} dataset.

ActivityNet Captions dataset~\cite{ActivityNet} is built on ActivityNet v1.3 which includes 20k YouTube videos. ActivityNet Captions is widely used in dense video captioning with annotated captions and timestamps of each sub-event.
The EgoSchema dataset~\cite{egoschema} contains over 5,000 video language understanding questions spanning over 250 hours of egocentric video data.
NExTVideo~\cite{NEXT} consists of 6,000 videos that are longer and richer in objects and interactions in daily life. 

We carefully select 1,438 videos (5.6k events) encompassing complicated causal relations and a wide range of scenarios as a new dataset: the MECD dataset, where 1,139 and 299 videos are randomly split for training and testing, respectively. {In our dataset, 1,106 videos are sourced from ActivityNet, 102 from EgoSchema, and 230 from NExTVideo.}

Specifically, each video in the MECD dataset contains 4 to 11 events, with a minimum of 2 premise events exhibiting causal relations with the last one. Fig.~\ref{fig: categories} presents the main categories and word clouds of video types.

{\noindent\textbf{Data Cleaning.}} As illustrated in ReXTime~\cite{rextime}, only around ten percent of event pairs in daily video datasets such as ActivityNet Caption~\cite{ActivityNet} are causal. 
Therefore, we further clean our dataset by excluding non-causal videos to guarantee the quality of our MECD dataset. 
We conducted data cleaning by five annotators with strict selection criteria: {If two or more annotators (out of five) label
a video as containing at most one causal chain, then the video will be excluded}, e.g. video that describes steps such as handcrafting. {A large portion of videos lacking causal relations were excluded—43.5\% from ActivityNet, 24.3\% from NExTVideo, and 74.2\% from EgoSchema.The results reveal that NExTVideo clips exhibit stronger causal coherence, while EgoSchema clips often lack it due to their ego-procedural nature.}

\noindent\textbf{Dataset Annotation.} \label{anna}The annotations of MECD dataset include 4 attributes. 
For videos sourced from the ActivityNet Captions dataset, The ``duration'', ``sentence'', and ``timestamps'' attributes in annotations remain the same as the original annotations for the dense video captioning task. 
{For videos sourced from EgoSchema and NExTVideo, we select samples that are drawn from Event-Bench's~\cite{eventbench} collection, where all clips have been verified to contain discrete events within standardized 30-second intervals. We then caption each event using the SOTA~\cite{videollama2} video captioning model Gemini-Pro~\cite{gemini}. Additionally, our annotators further refine the ``sentence'' and ``timestamps'' attributes to: (1) refine low-quality captions resulting from background clutter or ambiguous frames, and (2) improve event continuity that might be compromised by strict 30-second division.}

Specifically, in the context of our task, a new attribute, ``causality'', is introduced. This attribute represents the causal relations between all premise events $\{e_1, ..., e_{N-1}\}$ and the final event $e_N$.
To obtain this attribute, relations among events are first annotated using the GPT-4 API~\cite{gpt4}, and subsequently refined by five human annotators. Through a cross-annotation process~\cite{cross2,cross3}, ground truth causalities are determined by the majority of the annotators' causal relation choices, thus mitigating potential inaccuracies and subjective biases to a certain extent. 
{Besides, a Pearson correlation coefficient of 0.93 was achieved among the annotations, strongly indicating a high degree of objectivity in the annotation process. Our experiments also confirmed that the model's ability to learn causality is robust to minor variations in the annotations.}
Annotation examples of MECD are shown in Fig.~\ref{fig:sup_annotations2}, our MECD dataset is carefully annotated to support the challenging task proposed with complete premise information. 

{Furthermore, while the accuracy of deducing causality between the final event $e_N$ and its preceding events is a good general indicator, it doesn't fully capture the model's ability. 
To rigorously demonstrate the model's capacity to deduce the complete causal graph}, an additional attribute, ``complete causality'', is introduced for the test set. 
This attribute represents all causal relations between any two events, and is annotated and refined in the same way as the ``causality'' attribute. 
Specifically, ``complete causality'' is a list of length (N-1), where N is the number of events in the video, and the k-th item (length = (k+1)) in this complete causality list represents the causal relations between  $\{e_1, ..., e_{k+1}\}$ and $e_{k+2}$ as shown in Fig~\ref{fig:sup_annotations2}. 
{The task of discovering the ``complete causality'' is defined as \textbf{complete causal graph discovery}.}

{\noindent\textbf{Data Statistics.}} 
Our MECD dataset is primarily designed for causal reasoning in moderately long-duration videos, as illustrated in Fig.~\ref{fig: statistics} (a1) and (a2), the videos under consideration are predominantly those exceeding two minutes in duration, and all videos contain at least four distinct sub-events, events can be considered as complete actions with scenes in terms of our MECD task.

We also present the ratio of causal relations {within every video} in Fig.~\ref{fig: statistics} (b1), and the impact of events' positions on their causality in Fig.~\ref{fig: statistics} (b2). 
The proportion of events with or without causal relations is generally balanced, with the proportion of events without causal relations slightly higher. 
Moreover, by examining the last event as a case study, we investigated the correlation between causality and positional relations among events. 
Our findings indicate that the likelihood of adjacent events existing causal relations is higher, while the probability of other positions is comparable.

\section{Method}
\begin{figure*}[t]
\begin{center}
\includegraphics[width=0.9\textwidth]{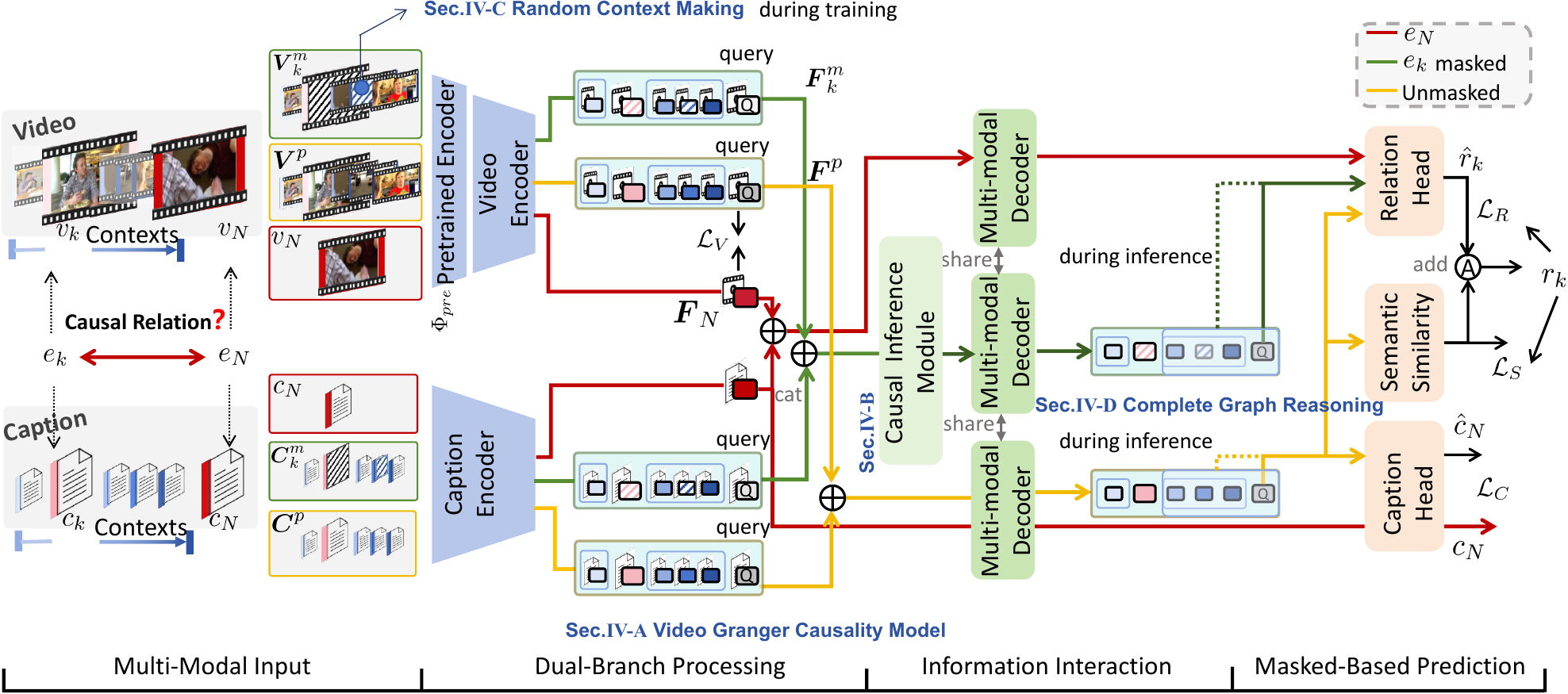}
\end{center}
\vspace{-5pt}
\caption{\textbf{Video Granger Causality Model.} Two data streams \{${\boldsymbol V}^p, {\boldsymbol C}^{p}$\} (with certain $e_k$) and  \{${\boldsymbol V}^m_k, {\boldsymbol C}^m_k$\} (without $e_k$) serve as input, video and text embeddings are concatenated after being separately embedded. During dual-branch processing, these two streams are encoded, and further multi-modal information interactions are conducted. The final mask-based prediction module conducts reasoning with two decoded prediction features and ground truth features embedded by \{${\boldsymbol v}_N, {\boldsymbol c}_N$\}. Further causal inference is conducted for mitigating causality confounding and illusory as in Sec.~\ref{causal}, and random context masking and non-regressive complete graph reasoning are introduced for robust and efficient reasoning as in Sec.~\ref{sec: context} and Sec.~\ref{sec: complete}}
\label{baseline}
\end{figure*}

In this section, we present our proposed Video Granger Causality Model (VGCM) to address Multi-Event Causal Discovery (MECD), as shown in Fig.~\ref{baseline}. 
This model establishes the global connections across all events, and deduces the causality of a premise event by comparing the output features when it is masked or not, under the concept of the \textit{Event Causality Test}. 
However, masking out an event may lead to the problem of confounding and illusion. 
{We further employ causal inference methods to mitigate confounding by compensating for or removing the effects of preceding or subsequent causal events. Meanwhile, during inference, the extra chain of thoughts and existence-only descriptions help alleviate potential illusions.}
Additionally, we extend the modeling of event-causal relations to the global level and utilize context chain causal reasoning to increase the interaction of global information.

\subsection{VGCM: Video Granger Causality Model}
\label{chapter:baseline}
Building upon the Granger Causality method introduced in~\cite{sup1,sup2,sup3}, our core motivation for constructing VGCM is \textit{Event Causality Test}: To compare the prediction of the last event using all the premise events with or without a certain event in it. If the results exhibit obvious divergence, it indicates that the current event is causally related to the result event.

We design VGCM to take in both the video clips and the captions to maximize information utilization. As illustrated in Fig.~\ref{baseline}, our proposed VGCM is a multi-modal transformer-based structure with a video encoder and caption encoder, and a multi-modal decoder with causal relation head to discover causal relations through the predicting process and the comparison of predicting results.

Based on this, we denote $\mathbb{E}^p$ as the set of all the \emph{premise} events $\mathbb{E}^p:=\mathbb{E} \setminus e_N $, and $\mathbb{E}^m_k:=\mathbb{E}^p \setminus e_k$ as the event set where the premise event $e_k$ ($k<N$) is masked. Notably, we mask the event $e_k$ by setting all zeros to its video clip $v_k$ and assign constant characters to the caption $c_k$. 

Following~\cite{VAR,dvc1,dvc2,dvc3}, we firstly pretrain a video encoder $\Phi_{pre}$ under an action recognition task to extract the features of the video clips. We essentially create two paths, one for the unmasked event set $\mathbb{E}^p$ (\textcolor{mycolor_orange}{orange path} in Fig.~\ref{baseline}) while the other for the set with one event (\textit{i.e.}, $e_k$) masked $\mathbb{E}^m_k$ (\textcolor{mycolor_green}{green path} in Fig.~\ref{baseline}). The video clips and captions are first separately encoded using $\text{Enc}_V$ and $\text{Enc}_C$ to obtain compact features, then their features are sent to a multi-modal decoder $\text{Dec}$ that shares weights for both paths to fuse the information. Afterward, several model heads are employed for feature comparison and loss measurement.
${\boldsymbol V}^p$ and ${\boldsymbol C}^p$ are the video clip and caption matrix concatenated from all premise events set $\mathbb{E}^p$, similarly, ${\boldsymbol V}^m_k$ and ${\boldsymbol C}^m_k$ are from $\mathbb{E}^m_k$. 
\begin{equation}
\begin{aligned}
{\boldsymbol F}^p&=\text{Enc}_V(\Phi_{pre}({\boldsymbol V}^{p})),\\
{\boldsymbol O}^p&=[\text{Dec}(\texttt{{Cat}}({\boldsymbol F}^p,\text{Enc}_C({\boldsymbol C}^{p}))]_{N-1},\\
{\boldsymbol F}^m_k&=\text{Enc}_V(\Phi_{pre}({\boldsymbol V}^m_k)),\\
{\boldsymbol O}^m_k&=[\text{Dec}(\texttt{{Cat}}({\boldsymbol F}^m_k,\text{Enc}_C({\boldsymbol C}^m_k))]_{N-1},
\label{embedding}
\end{aligned}
\end{equation}
where $\text{Enc}_V$ and {$\text{Enc}_C$} represent the encoder module of video clips and captions, respectively. $\text{Dec}$ is a multi-modal decoder that shares weights for both paths. $\texttt{{Cat}}$ indicates the concatenate operation, and $[-]_{N-1}$ indicates the $(N-1)$-th slice at dimension 0. ${\boldsymbol F}^p$ and ${\boldsymbol F}^m_k$ are features after encoding, and ${\boldsymbol O}^p$ and ${\boldsymbol O}^m_k$ are the output features, which are then used for comparison of difference. Incorporating both visual and linguistic representations, the decoder conducts cross-modal reasoning and leverages the context from the unmasked premise events to posit a meaningful representation of the most likely explanatory result event. 

Subsequently, the feature ${\boldsymbol O}^p$ deduced from the unmasked events is sent to the caption head for supervised event prediction. Additionally,  in order to compare the difference of the predictions, ${\boldsymbol O}^p, {\boldsymbol O}^m_k$ are directed to the relation head for reasoning. 
The result event $e_N$ is encoded the same way as $e_k$ to get feature ${\boldsymbol F}_N = \text{Enc}_V(\Phi_{pre}(v_{N}))$ and the output ${\boldsymbol O}_N = \text{Dec}(\texttt{{Cat}}({\boldsymbol F}_N, \text{Enc}_C({\boldsymbol C}_{N}))$, ${\boldsymbol O}_N$ is aggregated for reasoning (\textcolor{mycolor_red}{red path} in Fig.~\ref{baseline}). The relation head consists of a semantic query module and a self-enhancement module, where outputs are concatenated and then passed through the cross-reasoning layer $g_{r}$ for further interaction. Last but not least, the auxiliary similarity is measured between ${\boldsymbol O}^p$ and ${\boldsymbol O}^m_k$ as a supplement to the output information of the relation head.
After the reasoning process, the prediction output of the causal relation $\hat{r}_k$ can be represented by:
\begin{equation}
\begin{split}
\label{eq:hat_r_k}
\hat{r}_k = g_{r}(\texttt{{Cat}}(\text{$\Phi_{att}^C$}(\texttt{{Cat}}({\boldsymbol O}^m_k,{\boldsymbol O}_N),\texttt{{Cat}}({\boldsymbol O}^p,{\boldsymbol O}_N)), \\
\text{$\Phi_{att}^I$}(\texttt{{Cat}}({\boldsymbol O}^m_k,{\boldsymbol O}_N)))),
\end{split}
\end{equation}
where $\text{$\Phi_{att}^C$}$ represents cross-attention, $\text{$\Phi_{att}^I$}$ represents self-attention, $g_{r}$ represents linear layer. 
The training objective consists of two main directions as previously discussed: 

To reconstruct the textual and visual representation of the result event $e_N$, we introduce caption loss and reconstruction loss, respectively. Caption loss $\mathcal{L}_{C}$ ensures an accurate prediction of the result caption $\hat{c}_N$ given all the premise events $\mathbb{E}^p$. Simultaneously, visual reconstruction loss $\mathcal{L}_{V}$ forces the encoder to  ``imagine'' a representation of the result video clip $\hat{v}_N$ that better aligns with the original representation ${v_N}$. These losses allow the model to predict visual and textual representations that are close to the original representations, which better supports our method of inferring causal relations by comparing the results of the two-stream predictions.

For the objective of causal discovery, we introduce causal relation loss and an auxiliary semantics similarity loss. Causal relation loss $\mathcal{L}_{R}$ supervised the output relations $\hat{r}_k$. Meanwhile, the semantics similarity loss $\mathcal{L}_{S}$ is introduced to guarantee the semantics similarity of result event prediction under the presence or absence of a causal-relation-free premise event. The complete loss function is:
\begin{equation}
\begin{split}
\mathcal{L} = \mathcal{L}_{C}(c_N,\hat{c}_N) + \lambda_R \mathcal{L}_{R}(r_k,\hat{r}_k) + \lambda_V \mathcal{L}_{V}({\boldsymbol F}^p_N,{\boldsymbol F}_N) + \\
\lambda_S \text{sign}(r_k)\mathcal{L}_{S}(\boldsymbol O_{m}^k,{\boldsymbol O}^{p}),
\end{split}
\end{equation}
where $\lambda_R$, $\lambda_V$, and $\lambda_S$ are weights for trade off. $\mathcal{L}_{C}$ and $\mathcal{L}_{R}$ are the cross-entropy losses, $\mathcal{L}_{V}$ and $\mathcal{L}_{S}$ are the mse losses, ${\boldsymbol F}^p_N$ is the Nth slice of ${\boldsymbol F}^p$ indicating the feature of $e_N$.

Overall, the causal relation loss $\mathcal{L}_{R}$ provides direct supervision for output causal relations using standard Cross-Entropy loss, while $\mathcal{L}_{C}$, $\mathcal{L}_{V}$, and $\mathcal{L}_{S}$ aim to further enhance the causal discovery capability from different aspects. Specifically, $\mathcal{L}_{C}$ and $\mathcal{L}_{V}$ introduce auxiliary caption and reconstruction losses to facilitate event prediction, implemented by Label Smoothing loss and MSE loss respectively. They are incorporated because the Granger Causality Method determines whether an earlier event aids in predicting a subsequent one. $\mathcal{L}_{S}$, the similarity loss (implemented by InfoNCE), prompting causal relation discovery by comparing output causal feature similarity when a premise event is masked. The rationale behind this loss is that masking a non-causal event should result in a prediction of the result event similar to that of the unmasked stream.

\subsection{Causal Inference}
\label{causal}

In Sec.~\ref{chapter:baseline}, we employ the concept of Granger Causality to design our VGCM model under the principle of \textit{Event Causality Test} which may, however, introduce causality confounding and illusory.  Below we introduce these issues in detail, as well as how we manage to solve the problems. 

\textbf{Causality confounding} is a phenomenon where the original causal relations across events are impacted due to modification (\textit{i.e.}, masking) of some intermediate events (\textit{i.e.}, $e_k$).
Existing disentangled representation learning works~\cite{ica1,ica2} disentangled different attributes of a variable under strict assumptions but failed in disentangling different variables.

\begin{figure*}[t]
\begin{center}
\includegraphics[width=0.7\textwidth]{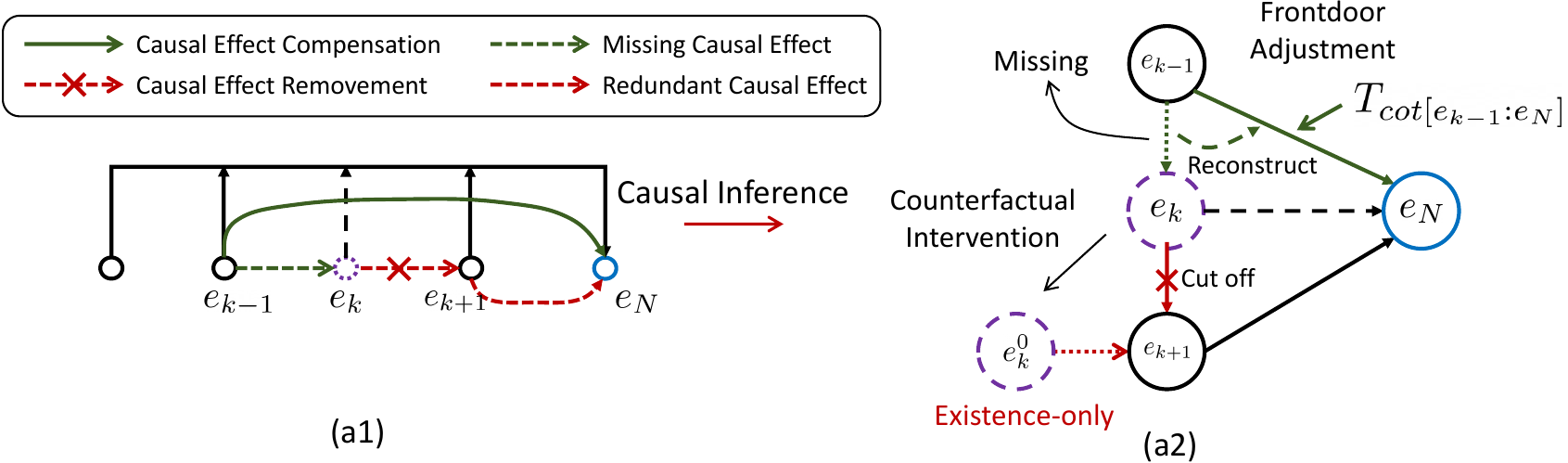}
\end{center}
\vspace{-10pt}
\caption{{\textbf{Causal Effect of the Adjacent Events and Causality Graph.}} (a1) shows the causality of $e_k$ analyzed, the causal effect in red needs to be mitigated because the causal effect residual remains through $e_{k+1}$. While the green needs to be compensated because the causal effect of $e_{k-1}$ is affected. (a2) shows the causal inference methods (Front-door Adjustment and Counterfactual Intervention) corresponding to the two causal effects.}
\label{causal_diagram}
\vspace{-5pt}
\end{figure*}

Specifically, when $e_k$ is masked for the comparison in VGCM, the causal relations between $e_k$'s adjacent events and the last event are impacted, leading to a confounding of causal effects. Notably, for brevity, we only employ $e_k$'s previous one event $e_{k-1}$ and its subsequent one event $e_{k+1}$ for analysis, but the same analysis also applies to all the previous or subsequent events. To be specific, there exist two distinct kinds of confounding when $e_k$ is absent:
\textbf{1)} Causal effects of 
$e_{k-1}$ to $e_N$ may be lost, as its connection to $e_N$ is built upon $e_k$, (\textcolor{mycolor_green}{green path} in Fig.~\ref{causal_diagram} (a1)).
\textbf{2)} Causal effects of $e_{k+1}$ to $e_N$ may be redundant, {as $e_k$ may be a necessary prior of $e_{k+1}$'s connection with $e_N$}, (\textcolor{mycolor_red}{red path} in Fig.~\ref{causal_diagram} (a1)).

\textbf{Illusory causality} is another issue that may lead to some spatial or temporal misunderstandings, including illusory temporal and existence causality. \textbf{1)} Illusory temporal causality is the situation that events could have tight temporal ordering, but they in fact have no causal relations. 
\textbf{2)} Additionally, illusory existence causality occurs when an object introduced in the premise event is a necessary condition for the result, but the premise event does not semantically serve as a reason.
Notably, we find that illusory in multi-event videos is much more significant than two independent events, which also tends to be exacerbated by causality confounding. 

Overall, \textbf{causality confounding} and \textbf{illusory causality} both bring difficulties for relation modeling of events in videos. Notably, these two issues are coupled in that \textbf{causality confounding} tends to exacerbate \textbf{illusory causality} by misallocating attention to temporal ordering and causal effect. Therefore, illusory causality can be partially relieved by solving the problem of causality confounding.

When considering taking the illusory causality, the chain of thoughts~\cite{cot1,cot2,cotsurvey} has been shown in LLMs to lead the model to logical thinking which is similar to human thought process, the chain of thoughts $T_{cot[e_{k-1}: e_N]}$ provides a step-by-step process of reasoning the $e_N$ from $e_{k-1}$. Specifically, $T_{cot[e_{k-1}: e_N]}$ is obtained using GPT-4 API~\cite{gpt4} by feeding it with $e_{k-1}$, $e_N$ along with a prompt asking it to provide the probable reasoning chain. We consider utilizing it in causal inference to eliminate the attention bias on temporal correlations introduced by non-causal temporal knowledge. 

Besides, as the illusory existence causality is caused by the objects' correlation between the events, we address this influence by keeping objects in the \textcolor{mycolor_green}{green path} in Fig.~\ref{baseline} the same as those in the \textcolor{mycolor_orange}{orange path}. We introduce an alternative event $e_k^0=\{v_k^0, c_k^0\}$ of $e_k$ to briefly recaps the objects in $e_k$. Specifically, $c_k^0$ is obtained using GPT-4 API~\cite{gpt4} by feeding it with $c_k$ along with a prompt asking it to extract the objects from $c_k$ and organize them as the sentence such as ``There are objects A, B and C.''. 
Consequently, we opt to employ $c_k^0$ to approximate $e_k^0$ in our VGCM model while omitting $v_k^0$, as $c_k^0$ is sufficient already to convey the information of objects. By providing $e_k^0$, all the necessary objects are still available in this path, thus effectively mitigating the illusory existence causality, facilitating the model to focus more on essential and causality-related semantic information.

To tackle the issues above, we introduce two causal inference methods: the front-door adjustment~\cite{causality} for the missing causal effect of $e_{k-1}$ and counterfactual inference~\cite{causality} for the redundant causal effect of $e_{k+1}$. Meanwhile, the chain of thoughts $T_{cot[e_{k-1}: e_N]}$ and the descriptions of existence $c_k^0$ are also provided to carefully address illusory causality, which in turn mitigates confounding.

We establish a causality graph in Fig.~\ref{causal_diagram} (a2) for an improved elaboration.
On masking $e_k$, the causality confounding that requires compensation {$\boldsymbol{F}^C$} or removement {$\boldsymbol{F}^R$} can be expressed as:
\begin{equation}
\begin{aligned}
\boldsymbol{F}^C &= P(e_N|e_k) - P(e_N|do(e_k)), \\
\boldsymbol{F}^R &= P(e_N|e_{k+1}) - P(e_N|do(e_{k+1})),
\label{front}
\end{aligned}
\end{equation}
 where $P(e_N|e_k)$ and $P(e_N|e_{k+1})$represents the process by which we predict $e_N$ from $e_k$ and $e_{k+1}$ in the \textcolor{mycolor_orange}{orange path} in Fig.~\ref{baseline}, and $do(\cdot)$ represents do-operation in causal inference~\cite{causalinference} that cuts off the causal relation between the event and its causes.

We aggregate the subsequent events $e_{k+1}$, the current event $e_k$ and the chain of thoughts $T_{cot[e_{k-1}: e_N]}$ using a linear layer $g_{do}$ for aggregation and the cross-attention and self-attention, according to the study in~\cite{linliangpami,hanwangzhang}, $P(e_N|do(e_k))$ can be implemented as:
\begin{equation}
\begin{split}
\label{eq:P_eN_do_ek}
P(e_N|do(e_k)) = g_{do}((\texttt{{Cat}}(\Phi_{att}^C(e_k,e_{k+1},e_{k+1}), \\
\Phi_{att}^I(e_k,e_k,e_k), \text{Enc}_c(T_{cot[e_{k-1}: e_N]}))),
\end{split}
\end{equation}
 Here, we re-use the cross-attention $\Phi_{att}^C$ and the self-attention $\Phi_{att}^I$ modules as in \eqref{eq:hat_r_k} to cut off the causal effect from $e_{k-1}$ to $e_k$ through do-operation, $e_k$ only interacts with subsequent events in predicting $e_N$. Then the missing causal effect $\boldsymbol{F}^C$ can be compensated since the causal-view operation and illusory temporal causality can be suppressed at the same time with the introduction of the chain of thoughts. Similarly, the redundant causal effect $\boldsymbol{F}^R$ can be removed by applying counterfactual intervention, then $P(e_N|do(e_{k+1}))$ can be represented by:
\begin{equation}
\label{eq:P_do_e_k+1}
P(e_N|do(e_{k+1})) = P(e_N|e_{k+1})[P(e_{k+1}|e_k)-P(e_{k+1}|e_{k}^0)],
\end{equation}
$P(e_N|do(e_{k+1}))$ effectively cuts off the redundant causal effect between $e_{k+1}$ and $e_N$ for the reason that the causes of $e_{k+1}$ are replaced with counterfactual description $e_k^0$, then the illusory existence causality can be suppressed simultaneously.

To refine the originally decoded feature ${\boldsymbol O}^m_k$ from the path with premise events masking:
\begin{equation}
 {\boldsymbol O}_k^{\prime m} = {\boldsymbol O}^m_k - \text{Dec}(\boldsymbol{F}^C) + \text{Dec}(\boldsymbol{F}^R),
\end{equation}
where ${\boldsymbol O}_k^{\prime m}$ is the refined feature that replaces ${\boldsymbol O}^m_k$ for further deduction of the model. With the refinement feature ${\boldsymbol O}_k^{\prime m}$, our VGCM model effectively compensates the connections between $e_{k-1}$ and $e_{N}$ that were originally lost due to the removal of $e_k$, and effectively removes the redundant causal effect between $e_{k+1}$ and $e_{N}$ as well.

\subsection{Context Chain Reasoning}
\label{sec: context}
As discussed above, when conducting the \textit{Event Causality Test}, a certain event $e_k$ is masked, and the prediction result of the last event $e_N$ utilizing all the premise events with or without $e_k$ is compared.

Although the \textit{Event Causality Test} complies with the Granger Causality methodology, when applied to event-level reasoning in videos, it slightly overlooks the contextual relations between different events within the same video. 
This overlook is disadvantageous to the model's reasoning ability according to the research in ~\cite{granger_event_context,granger_event_context2}, particularly in inferring the intricate causal patterns of the entire causal graph. 

\begin{figure}[t]
\begin{center}
\includegraphics[width=0.48\textwidth]{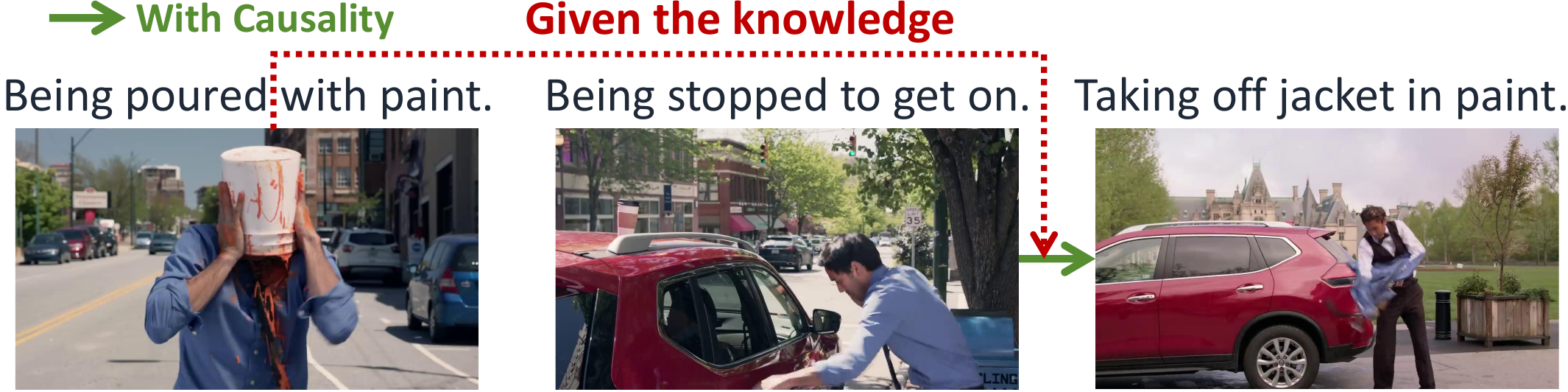}
\end{center}
\caption{{\textbf{An Example of Context Chain Reasoning.}} Given the knowledge that paint is stuck on the clothes, it is easier to infer the causal relation between being prohibited from taking the car and taking off one's jacket.}
\label{context_event}
\vspace{-5pt}
\end{figure}

In reality, there are numerous instances where partial causes, when combined, contribute to the formation of the complete cause leading to the result as shown in Fig.~\ref{context_event}. 
Being prevented from getting into the car and taking off a jacket appears to have no causal relation. However, when considering prior knowledge that the jacket is covered in paint, this context aids in making a correct judgment through causal reasoning. Thus, we employ context chain reasoning to fully enhance the model's ability to leverage contextual events better. 
Therefore, we conduct the \textit{Event Causality Test} by considering several premise events simultaneously during the training phase to enhance the context reasoning ability. 

Specifically, during the training phase, we adopt a strategy of concurrently masking multiple events, $e_k = \{e_{k0}, ..., e_{ki}\}$, (where $0<i<(n-1)$), to enhance the model's consideration of contextual events during reasoning.
 The ground truth causal relation is represented by the labels of the multiple events $e_k = \{e_{k0}, ..., e_{ki}\}$ that are masked simultaneously, which are combined using an OR operation:~$r_k = \bigvee_{j=0}^{i} r_{kj}$. 

Furthermore, given that causal data in real-world events is relatively scarce~\cite{xiaoding}, the imbalance in the data resulting from the OR operation of the labels can mitigate the disadvantage of classification bias on model training~\cite{longtail}, further enhancing the model's generalization in judging causal relations.

\subsection{Non-regressive Complete Graph Reasoning}
\label{sec: complete}
Through the \textit{Event Causality Test} introduced in the previous sections, we can infer the causal chains between all premise events and the final event $e_N$. 
To further infer the complete event-level causal graph during the inference phase, a straightforward method treats each event as a result event and conducts \textit{Event Causality Test} using a regressive approach~\cite{granger_app}, ultimately synthesizing a complete causal graph. 
However, this regressive approach is highly inefficient because inferring a causal graph for a video with $N$ events requires the model to perform forward propagation $(N-1)(N-2)/2$ times. 
This quadratic time complexity is suboptimal, especially given that our dataset contains an average of more than five events per video. 
To tackle this, we employ the following efficient non-regressive complete graph reasoning method during inference.

To maintain generality, we focus on the task of inferring the causal relation $\hat{r}_{[i:j]}$ between any of the $i$-th and $j$-th events, (where $1 \leq i < j \leq (N-1)$). 
As the \textit{Event Causality Test} introduced above, the mask-based reasoning process of relation head involves two prediction features under different conditions for comparison: ${\boldsymbol O}^p$ and ${\boldsymbol O}^m_i$, which correspond to the $j$-th slice of the decoder output features defined in Eq.~\ref{embedding}. 
Two output features of event ${\boldsymbol O}^p$ (when $e_i$ is not masked) and ${\boldsymbol O}^m_i$ (when $e_i$ is masked) can be represented by:
\begin{equation}
\begin{aligned}
{\boldsymbol O}^p &= [\text{Dec}(\texttt{{Cat}}({\boldsymbol F}^p_i, \text{Enc}_C({\boldsymbol C}^{p}))]_{j},\\
{\boldsymbol O}^m_i &=[\text{Dec} (\texttt{{Cat}}({\boldsymbol F}^m_i, \text{Enc}_C({\boldsymbol C}^m_k))]_{j},
\label{embedding_whole}
\end{aligned}
\end{equation}
where $k$ equals $i$, as $k$ iterates from $1$ to $N-1$ during inference of causal relations with $e_N$, no additional forward propagation is required. 
Our approximation method specifically involves utilizing the features ${\boldsymbol O}^p$ and ${\boldsymbol O}^m_i$ without additional masking of $e_j$, following the approach outlined in ~\cite{granger_app}.
The ground truth result event $e_j$ is encoded in the same manner as $e_i$, yielding the encoded feature ${\boldsymbol F}_j = \text{Enc}_V(\Phi_{pre}(v_{j}))$ and the output feature ${\boldsymbol O}_j = \text{Dec}(\texttt{{Cat}}({\boldsymbol F}_j, \text{Enc}_C({\boldsymbol C}_{j}))$. Following a similar mask-based reasoning as Eq.~\ref{eq:hat_r_k}, after reasoning, the output causal relation $\hat{r}_{[i:j]}$ can be represented by:
\begin{equation}
\begin{split}
\label{eq:hat_r_k_whole}
\hat{r}_{[i:j]} = g_{r}(\texttt{{Cat}}(\text{$\Phi_{att}^C$}(\texttt{{Cat}}({\boldsymbol O}^m_i,{\boldsymbol O}_j),\texttt{{Cat}}({\boldsymbol O}^p,{\boldsymbol O}_j)), \\
\text{$\Phi_{att}^I$}(\texttt{{Cat}}({\boldsymbol O}^m_i,{\boldsymbol O}_j)))),
\end{split}
\end{equation}
As a result, just like inferring causal chains only related to the result event $e_N$, where inference requires just $(N-1)$ forward propagation, this non-regressive complete graph reasoning method equips the VGCM with rapid inference speed.

\section{Experiments}
\subsection{Implementation details}

\begin{figure}[t]
\begin{center}
\includegraphics[width=0.48\textwidth]{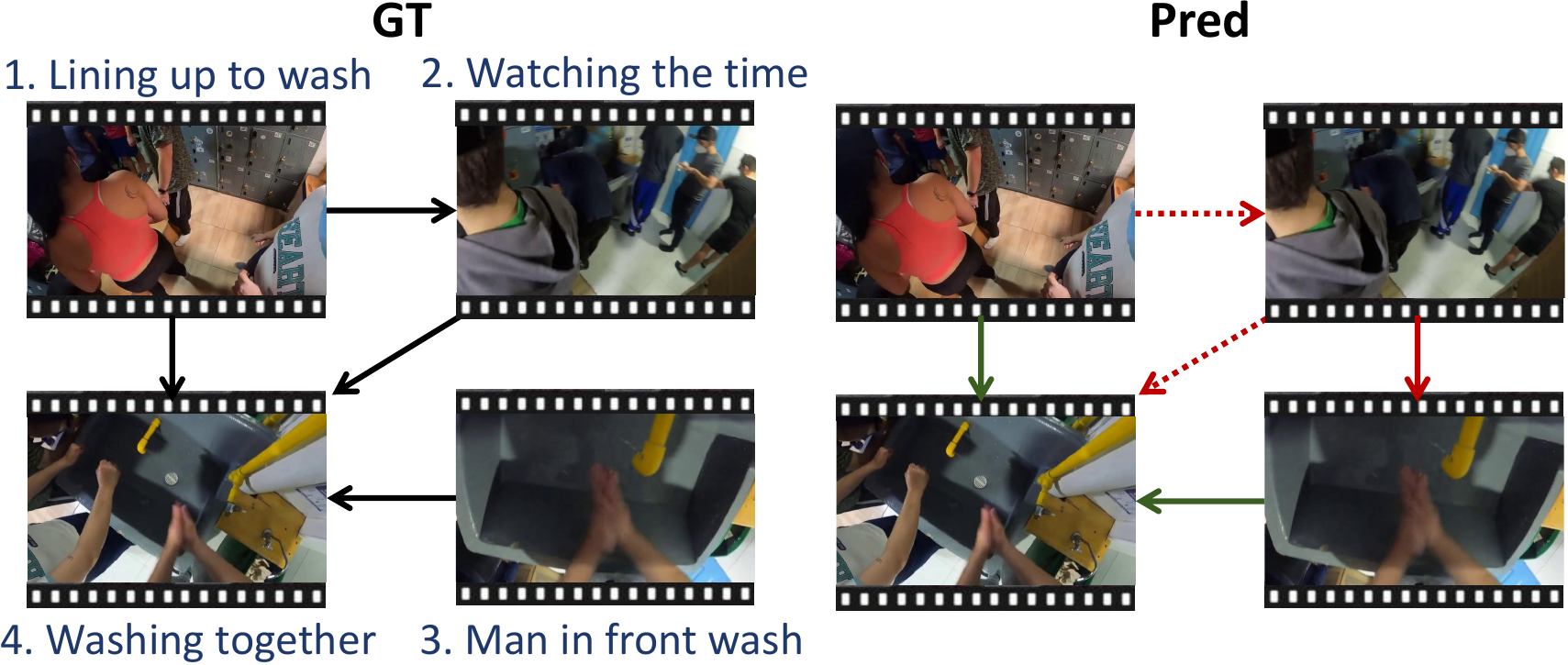}
\end{center}
\vspace{-5pt}
\caption{{\textbf{An Example of the Evaluation of SHD.} The figure illustrates that the prediction of the complete causal graph on the right contains two missing edges and one extra edge. Consequently, the total number of incorrectly predicted causal relation edges (SHD value) is 3.}}
\label{hamming}
\vspace{-5pt}
\end{figure}

Our encoder, $\text{Enc}_V$, $\text{Enc}_C$, and multi-modal video decoder, $\text{Dec}$, are built upon Videobert~\cite{videobert}, a joint model for video and language representation learning. Our model contains only 144M parameters, significantly smaller than 7B VideoLLMs.
$\lambda_C$, $\lambda_R$, $\lambda_V$, $\lambda_S$ are set to be 1.0, 4.0, 0.25, 0.05. Maximum input lengths of the caption, the chain of thoughts, and the existence-only descriptions are set to 50. 

During the pre-training process, visual features of each video are extracted using a ResNet-200 model~\cite{resnet} pre-trained on the ActivityNet dataset~\cite{caba2015activitynet_ar} under the action recognition task, following the approach in~\cite{dvc1,dvc2,dvc3}. 
Given the potential benefits of prior domain knowledge for the Granger Causality Causal method~\cite{overview}, we pre-trained our model for the dense video captioning task on a 3.1k-video dataset from the ActivityNet Captioning dataset~\cite{ActivityNet}, with each video sample containing more than four events. 


\noindent \textbf{Comparisons:}
We compare VGCM with baseline~\cite{videobert} and multi-modal base models such as CLIP-L~\cite{clip}, and a representative video reasoning model VAR~\cite{VAR}. 
Besides, we also conduct experiments on powerful LLM, including Mixtral-8x22B-Instruct~\cite{mixtral}, GPT-4~\cite{gpt4}, Gemini-Pro~\cite{gemini}, and so on. 
Additionally, ImageLLMs and VideoLLMs utilized for comparison include widely accepted GPT4-o~\cite{gpt4}, Gemini-Pro~\cite{gemini}, VideoLLaVA~\cite{videoLLaVA}, VideoChat2~\cite{mvbench}, {Long-VideoLLM LongVA~\cite{longva}, Qwen2.5-VL~\cite{qwen25vl},} and so on. 

In the setting of few-shot learning~(In-Context Learning), LLMs and ImageLLMs are evaluated, following the principle utilized in causal discovery tasks~\cite{few1, few2, few3} in NLP. 
To more comprehensively assess the capacity for causal reasoning and to leverage the benefits of open-source models, the performance of VideoLLMs and all multi-modal base models is assessed under a strong fine-tuning paradigm. 
For all VideoLLMs, the vision-language projector was fully fine-tuned using Lora on the training set of the MECD dataset. Similarly, for multi-modal base models, the vision and text encoders are frozen, and a causal relation classifier, composed of the MLP layers and sigmoid function, is fully tuned.

\begin{table*}[t]
    \centering
        \caption {\textbf{Main Results.} VGCM reaches SOTA performance on causal chains and complete causal graph discovery. {``w/o CR'' indicates without context chain reasoning.}}
    \resizebox{0.8\textwidth}{!}{
      \setlength{\tabcolsep}{3.5mm}{
      \begin{tabular}{ccccccc}
        \toprule
        Paradigm &Method &SHD~$\downarrow$ &Neg~$\uparrow$ &Pos~$\uparrow$ &Acc~$\uparrow$ \\
        \midrule
        \multirow{2}{*}{Random} 
         & Guess all causal. &6.95 &0.00 &100.00 &42.39 \\
        & Guess all non-causal. &5.36 &100.00 &0.00 &57.61 \\
        \midrule
        & \textit{Open-Source LLM}\\
        \multirow{18}{*}{$\text{In-context}$}
         
        

        & $\text{DeepSeek-Coder}$~\cite{deepseekcoder} &5.11 &67.27 &55.01 &61.97 \\

        & $\text{Mixtral-8x22B}$~\cite{mixtral} &\cellcolor{green!5}4.88 &64.54 &\textbf{65.60} &64.78  \\
        
        & $\text{Qwen-Max-0428}$~\cite{qwen} &4.95 &69.28 &58.57 &\cellcolor{green!5}64.80 \\

        \cmidrule{2-7}
        &\textit{Proprietary LLM}\\
        &   $\text{Gemini-1.5-Pro}$~\cite{gemini}  &4.75 &71.96 &57.32 &64.80\\
 
        &  $\text{GPT-4-0613}$~\cite{gpt4} &\cellcolor{green!10}\textbf{4.72} &\textbf{74.07} &56.80 &\cellcolor{green!10}\textbf{64.88}\\
        
        \cmidrule{2-7}
        &\textit{Proprietary ImageLLM}\\
        & Gemini-1.5-Pro~\cite{gemini} &4.70 &68.97 &\textbf{62.52} &65.15 \\
         
        & GPT-4o~\cite{gpt4} &\cellcolor{green!10}{\bf{4.69}} &\textbf{70.33} &61.53 &\cellcolor{green!10}\textbf{65.51} \\
    
        \cmidrule{2-7}
        & \textit{Open-Source VideoLLM}\\
        
        & MiniGPT-4~\cite{videochat} &5.12 &62.71 &51.92 & 58.29 \\
         
        & MiniGPT4-Video~\cite{minigpt4} &5.00 &65.72  &53.85 &59.88 \\
        
        & VideoLLaVA~\cite{videoLLaVA}  &4.95 &64.65 &\bf{57.46} & 62.78 \\

        &{$\text{LongVA}$}~\cite{longva} &\cellcolor{green!5}{4.82}  &{66.93}  &{56.10}  &{62.84} \\
                
        & PLLaVA~\cite{pllava}&4.86 &69.14 &55.88 & 63.00\\

        &{$\text{Qwen2.5-VL}$}~\cite{qwen25vl} &\cellcolor{green!5}{4.82}  &{65.37}  &{60.75}  &{63.59} \\
                
        & VideoChat2~\cite{mvbench} &\cellcolor{green!10}\textbf{4.75} &\textbf{70.63} &55.57 &\cellcolor{green!10}\textbf{63.85}\\

        \midrule
        & \textit{Open-Source Basic Multi-modal model}\\
        \multirow{13}{*}{$\text{Fine-tuning}$} 

        &$\text{Videobert}$~\cite{videobert} &4.95  &62.33  &57.82  &60.92 \\
        
        & $\text{VAR}$ ~\cite{VAR} &4.86 &59.79 &63.44 & 61.46 \\
        
        & \text{CLIP (ViT-L)}~\cite{clip} &{4.77} &63.88 &61.54 &62.87  \\
        
        & \text{SIGLIP}~\cite{siglip} &4.80 &\bf{64.13}  &{65.39}  &{64.85}   \\

        & {\text{BLIP-2 (ViT-L)}}~\cite{BLIP2} &{\bf{4.75}} &{63.75}  &{\bf{66.90}}  &{\bf{65.54}}  \\
        
        \cmidrule{2-7}
        & \textit{Open-Source VideoLLM}\\
       & PLLaVA~\cite{pllava} &4.74 &69.78 &63.35 & 65.71\\
        
       & VideoLLaVA~\cite{videoLLaVA} 
        &4.73 &66.69 &\bf{67.31} & 67.12 \\
        
        &{$\text{Qwen2.5-VL}$}~\cite{qwen25vl} &\cellcolor{green!5}{4.69}  &{69.37}  &{67.15}  &{68.06} \\
        
       & VideoChat2~\cite{mvbench} &\cellcolor{green!10}\bf{4.68}  &\bf{70.03}  &65.86  &\cellcolor{green!10}\bf{68.58} \\

       \cmidrule{2-7}
       & \textit{Ours} \\
        
       & \textbf{VGCM~(w/o CR)} &4.19 &\bf{76.58} &63.89 & 71.20\\
        
       & \textbf{\underline{VGCM}} & \cellcolor{green!25}\textbf{\underline{3.94}} &73.01 &\bf{68.63} & \cellcolor{green!25}\textbf{\underline{71.28}} \\
        
       \midrule
        
        \multirow{3}{*}{Human} &Only Textual Input &2.23 &87.47 &79.66 &83.50 \\
        
       & Only Visual Input &2.09 &88.09 &85.72 &87.02 \\
        
       & Complete Input &2.05 &88.32 &85.66 &87.19 \\
        \bottomrule
      \end{tabular}
    }}
    \label{tab:mainresults}
    \vspace{-10pt}
\end{table*}

\noindent \textbf{Metrics:}
Our model has two output formats: one that outputs {causal chains related to the result event}, focusing on the model's multi-cause reasoning ability {(\textbf{towards causal chains discovery task})}, and another that outputs a complete {causal graph of all events} {(\textbf{towards complete causal graph discovery task})}, ultimately reflecting the generalization ability in causal reasoning. 
Therefore, we designed two metrics to evaluate these aspects respectively.

We evaluate the model's causal discovery capability by utilizing the top-1 accuracy of the output causal relation chains related to the final result event. 
Following the approach in natural language processing~\cite{UniCE, xiaoding}, we further introduce the ``Neg'' metric to represent the accuracy when the model predicts no causal relation and the ``Pos'' metric to represent the accuracy when the model predicts the existence of a causal relation. 
This allows for a more nuanced analysis of whether the model tends to misjudge causal or non-causal relations.

In addition to the primary metrics, we introduce Structural Hamming Distance (SHD)~\cite{SHD1, SHD2} as a supplementary metric to evaluate the model's generalization ability in causal reasoning. 
SHD measures the degree of matching between comprehensive causal graphs by summing the number of incorrect causal relations. 
In the MECD test set, the average number of causal relations in video causal graphs is 12.31, and a lower Ave SHD value indicates better performance. 
An example of an evaluation of SHD is illustrated in Fig.~\ref{hamming}.

\subsection{Main Results}
As shown in Tab.~\ref{tab:mainresults}, our proposed VGCM achieves state-of-the-art results in causal chains and causal graph reasoning tasks, with an accuracy of 71.28\%. and an average SHD of 3.94.  
This surpasses existing MLLMs, including PLLaVA~\cite{pllava} and VideoChat2~\cite{mvbench}, as well as proprietary models such as Gemini-Pro~\cite{gemini} and GPT-4o~\cite{gpt4}. 
Specifically, VGCM demonstrates improvements of over 0.74 in SHD and 2.70\% in accuracy compared to the closest model, which exhibits reasoning capabilities most similar to human-level performance.

In the following \textbf{a) to d)}, we analyze the performances of VGCM, LLMs, VideoLLMs, and human reasoning.
Furthermore, we perform an illusory test to assess the hallucination tendencies.\\

\begin{figure}[t]
\begin{center}
\includegraphics[width=0.48\textwidth]{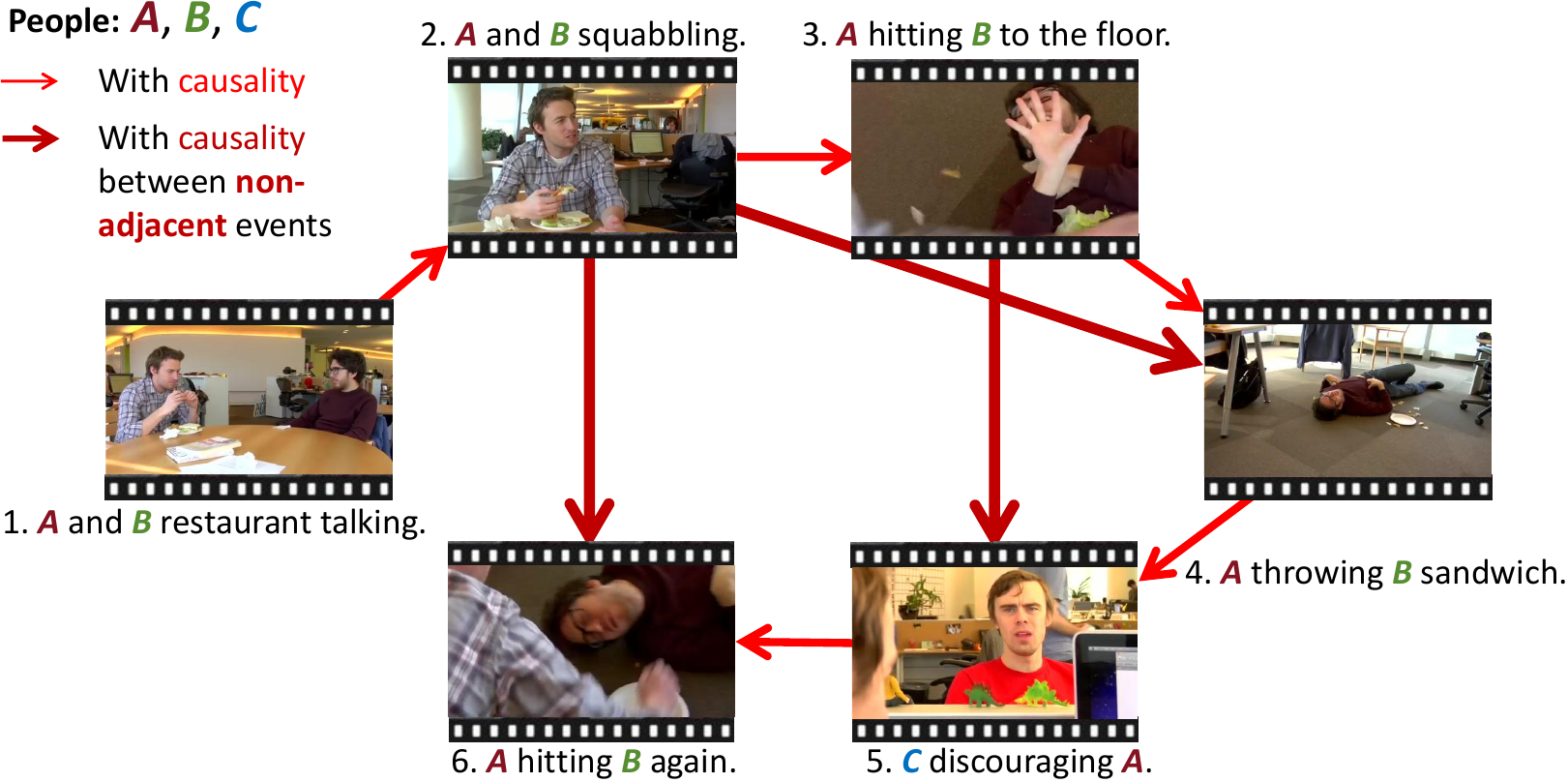}
\end{center}
\caption{\textbf{Complete Causal Graph.} The red arrows between events indicate causality, pointing from the cause to the result. Moreover, Events begin at the upper left corner and are sorted in a clockwise direction, with the causal relations between non-adjacent events being represented by thicker arrows.}
\label{fig: complete}
\end{figure}

\begin{table}[t]
      \centering
        \caption{\textbf{Inference Speed.} VGCM is 3-6 times faster than all VideoLLMs while slightly slower than the baseline. {w/o NR indicates without non-regressive complete graph reasoning.}}
        \resizebox{0.48\textwidth}{!}{
  \setlength{\tabcolsep}{2.0mm}
  \begin{tabular}{cc}
    \toprule
    Model &Inference Speed~(seconds/sample) \\
    \midrule
     Videobert~\cite{videobert} &0.70 \\
     \textbf{\underline{Our VGCM}} &\textbf{\underline{0.76}}\\
     CLIP~(ViT-B/32)~\cite{clip} &0.79 \\
     PLLaVA~\cite{pllava} &1.89 \\
     VideoLLaVA~\cite{videoLLaVA} &2.12\\
     VideoChat2~\cite{mvbench} &2.96\\
     {VGCM (w/o NR)} &{3.39}\\
     MiniGPT4-Video~\cite{minigpt4} &3.98\\
    \bottomrule
\label{speed}
  \end{tabular}
}
\vspace{-10pt}
\end{table}

\noindent\textbf{a) Our Performances.} As shown in Tab.~\ref{tab:mainresults}, our VGCM outperforming the SOTA Proprietary LLM GPT-4~\cite{gpt4}, ImageLLM GPT4-o~\cite{gpt4}, open-source basic multi-modal model SIGLIP~\cite{siglip}, VideoLLM VideoChat2~\cite{mvbench} by 6.40\%, 5.77\%, 6.43\%, and 2.70\%, respectively. 

Given that the Neg metric exceeds the Pos metric, our model initially shows a propensity to misclassify non-causal event pairs as causal. 
However, this issue is significantly alleviated following context chain reasoning, indicating that our model reduces the incidence of misjudged causes by enhancing the representation of contextual event relations.

Additionally, our VGCM further highlights its advantages in terms of complete causal graph reasoning, particularly without additional data supervision annotations.
VGCM achieves performance with only an average of 3.94 false causal relations for a complete causal graph. An example is shown in Fig.~\ref{fig: complete}.

As illustrated in Tab.~\ref{speed}, we have assessed the inference speed of various models, with our VGCM achieving a swift 0.76 seconds per sample. 
{VGCM with non-regressive complete graph reasoning incurs an overhead of only 8.57\% over the Videobert~\cite{videobert} baseline. 
VGCM's inference speed is 3 to 6 times faster than that of all VideoLLMs. This shows that VGCM with the non-regressive approach achieves both precise and efficient reasoning capabilities.} \\

\noindent\textbf{b) LLM Performances.} 
\begin{figure}[t]
\begin{center}
\includegraphics[width=0.48\textwidth]{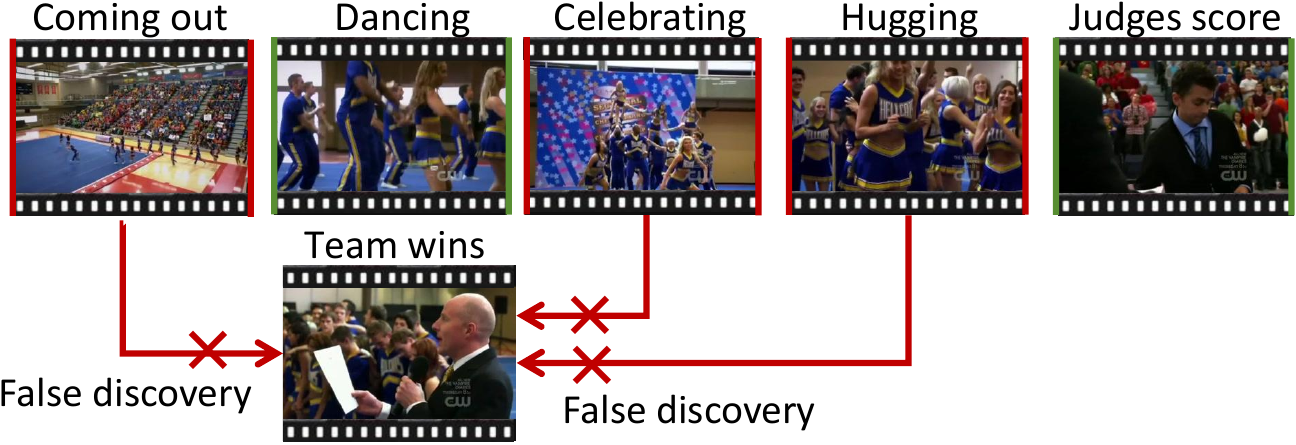}
\end{center}
\vspace{-5pt}
\caption{\textbf{Failure Abduction Examples of GPT-4.} Many failures of GPT's causal reasoning are due to confusion with illusions and the conflation of subjective emotions with objective laws.}
\label{fig:failure}
\end{figure}
\begin{figure}[t]
\begin{center}
\includegraphics[width=0.48\textwidth]{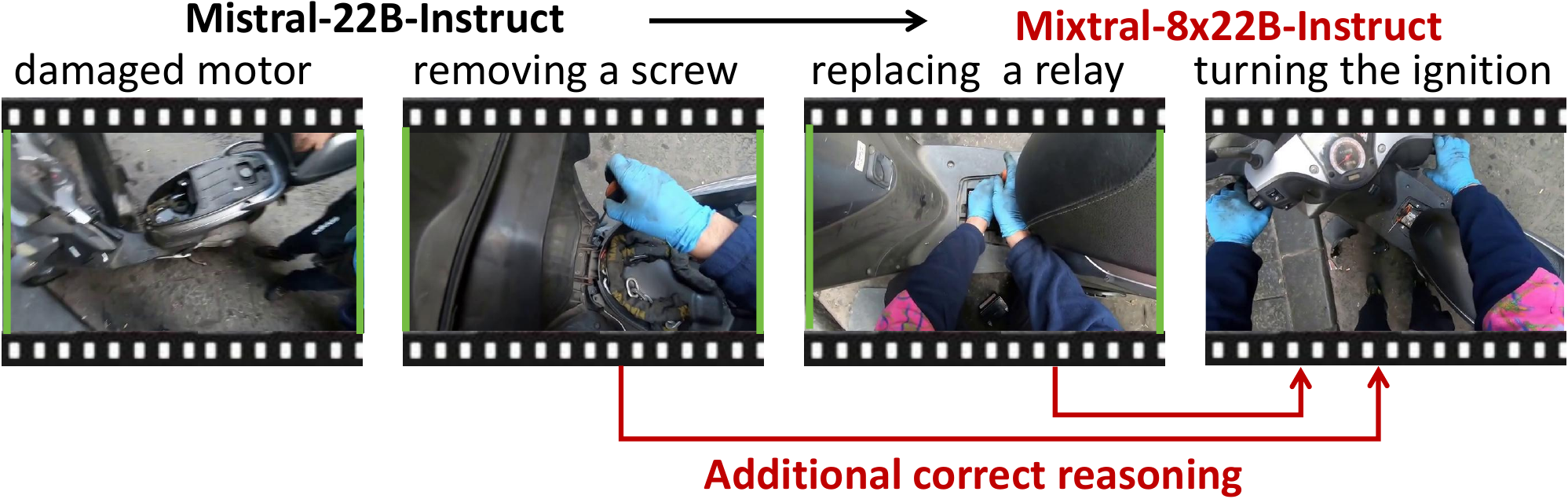}
\end{center}
\vspace{-5pt}
\caption{\textbf{{Comparison between Output Causal Chains of Mistral-22B-Instruct and Mixtral-8x22B-Instruct.}} LLM is facilitated with knowledge in a specific field of electrical appliances after utilizing a large-scale Mixture of Experts.}
\label{fig:expert}
\vspace{-10pt}
\end{figure}
As shown in Tab.~\ref{tab:mainresults}, Proprietary LLMs GPT-4~\cite{gpt4} and Gemini-Pro~\cite{gemini} have demonstrated the best performances among all LLMs. 
However, they are still limited by the influence of hallucinations and the conflation of subjective emotions with objective laws. 
As presented in Fig.~\ref{fig:failure}, GPT-4 incorrectly infers that all premise events have a causal relation with the result event of the team winning. 

Moreover, the Neg metric generally surpasses the Pos, which may be attributed to the training corpus's frequent inclusion of data instances where two events are causally linked, alongside a scarcity of examples where events are not causally related~\cite{xiaoding}. 
Meanwhile, the balanced performance of Mixtral-8x22B-Instruct~\cite{mixtral} may benefit from the use of a large-scale mixed expert model. 
As illustrated in Fig.~\ref{fig:expert}, after utilizing the MoE, Mixtral-22B~\cite{mixtral} successfully infers the causal knowledge involved in repairing electrical appliances.\\


\noindent\textbf{c) VideoLLM Performances.} 
As shown in Tab.~\ref{tab:mainresults}, under the paradigm of In-Context Learning, models such as VideoChat2~\cite{mvbench} and PLLaVA~\cite{pllava} have demonstrated superior performance, likely due to the inclusion of Next-QA~\cite{NEXT} and CLEVRER~\cite{CLEVRER} datasets in their pre-training data.
Furthermore, under the paradigm of Fine-tuning, the performances are enhanced, though they still fall short of the performance of our VGCM. {Moreover, under more frames of input, Long-VideoLLMs~(LongVA, Qwen2.5-VL) do not outperform general VideoLLMs. However, their relatively better SHD performance indicates that long-sequence perception enhances the ability to reason about overall causal relations.}

\begin{figure}[t]
\begin{center}
\includegraphics[width=0.48\textwidth]{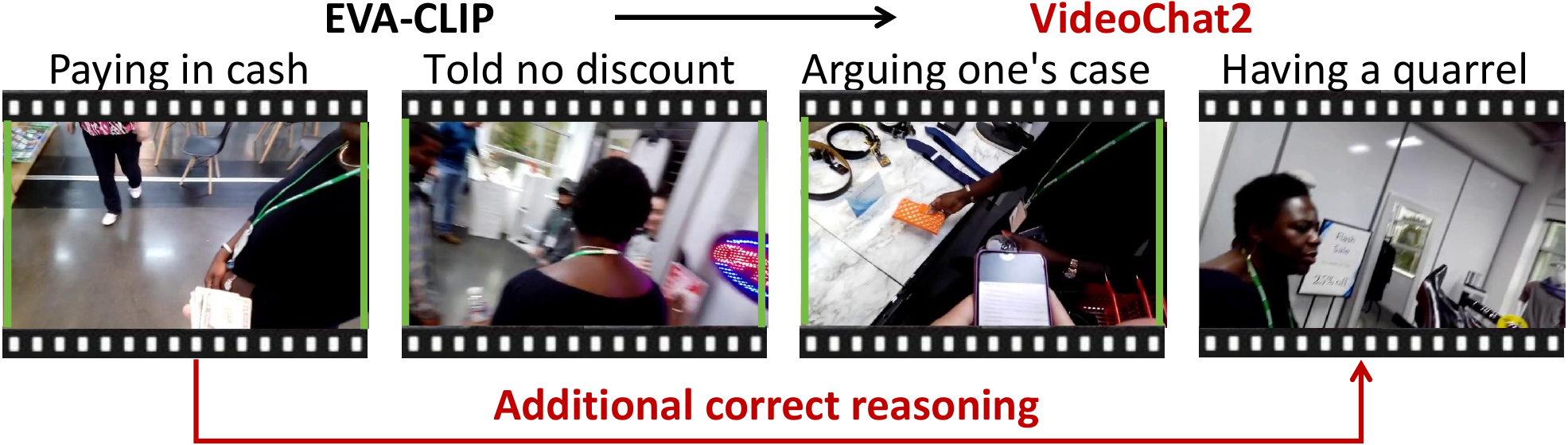}
\end{center}
\caption{\textbf{{Comparison between Output Causal Chains of EVA-CLIP and VideoChat2.}} VideoChat2 could be attributed to utilizing spatiotemporal cues to infer causal relations.}
\label{fig:chat2}
\vspace{-10pt}
\end{figure}

Besides, the performance of VideoLLMs suggests a reduction in the disparity between the Pos and Neg metrics, which may be attributed to the mitigation of hallucinations and the decrease in the influence of caption ambiguity due to adequate visual information. 
Furthermore, the enhancement in the Pos metric for VideoChat2~\cite{mvbench} could be attributed to the utilization of spatiotemporal cues to infer causal relations~\cite{mvbench}. 
As shown in Fig.~\ref{fig:chat2}, after utilizing spatiotemporal cues, compared to the same encoders with MLP structure model EVA-CLIP~\cite{evaclip}, VideoChat2~\cite{mvbench} successfully discovered the long-term causal relations between the initial reason event ``paying in cash'' and the result event ``having a quarrel'', reasoning leads to the ultimate source of results.

{The reasons for the 
performance gap between VideoLLMs and VGCM are summarized here: (1) VGCM is tailored for causal reasoning, while general VideoLLMs lack causal understanding due to insufficient exposure during pretraining; (2) VGCM models bidirectional dependencies using the Granger Causality framework, enabling robust causal inference, while the autoregressive nature of VideoLLMs enforces an autoregressive bias that limits their ability to capture global event relations.}\\

\begin{figure}[t]
\begin{center}
\includegraphics[width=0.48\textwidth]{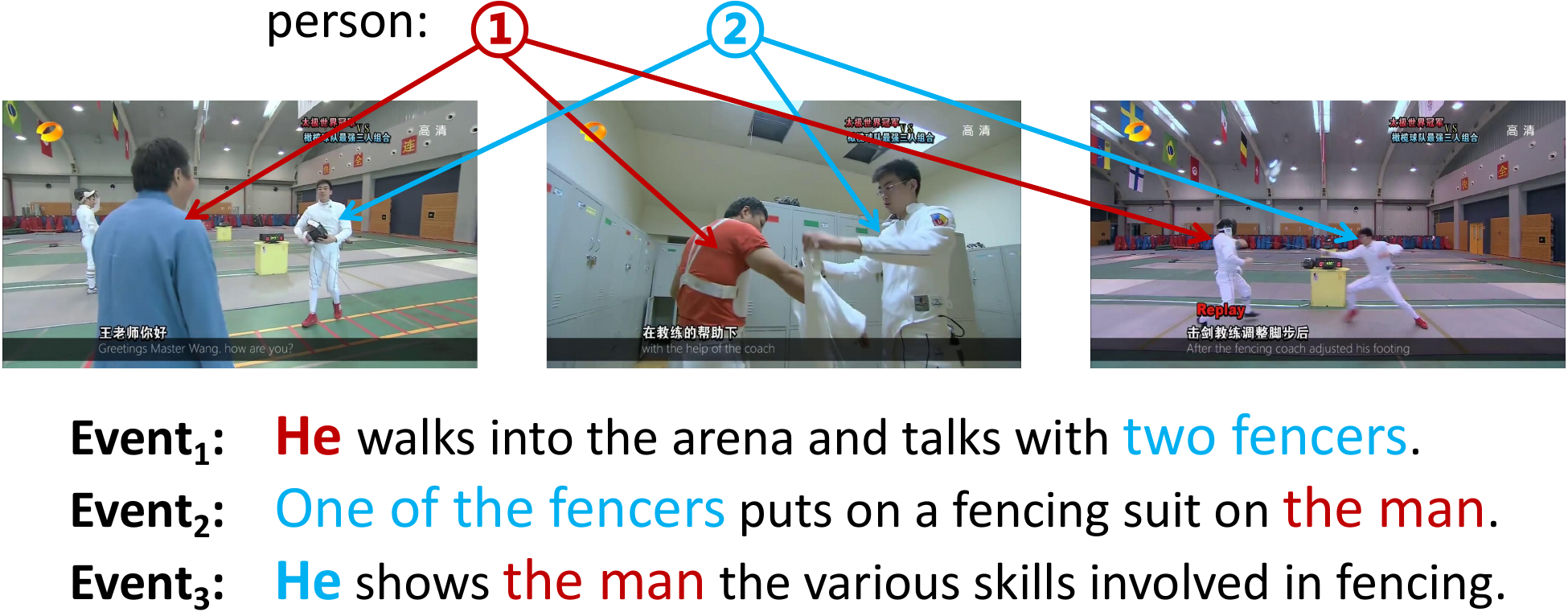}
\end{center}
\caption{\textbf{Examples of Confusion due to Unclear Caption.} The red arrows and text descriptions correspond to Person 1, while the blue ones correspond to Person 2. Evidently, the pronoun ``He'' successively refers to different individuals.}
\label{man}
\end{figure}
\noindent\textbf{d) Human Performances.} 
As shown in Tab.~\ref{tab:mainresults}, the average performance of ten volunteers reaches 87.19\%.
When only inputting one modality of information, higher accuracy is achieved when visual information is the sole input. 
We posit that this may be attributed to ambiguous pronouns can affect the judgment of their referents. An example of caption confusion is shown in Fig.~\ref{man}, however, this problem can be released with additional visual input.\\

\begin{table}[t]
    \centering
        \caption {\textbf{Illusory Causality Test.} \dag~indicates with fine-tuning paradigm, while * indicates with In-Context Learning paradigm, the Change reflects the accuracy variance compared with the MECD test set. LLMs and ImageLLMs suffer serious illusions, however, VGCM and VideoLLMs do not suffer.}
    \resizebox{0.48\textwidth}{!}{
      \setlength{\tabcolsep}{2.0mm}{
      \begin{tabular}{llcc}
        \toprule
        &Method & Accuracy & Change\\
        \midrule
        \multirow{5}{*}{$\text{LLM}^*$}  
        & $\text{GLM-4 (0520)}$~\cite{zhipu} & 54.96 &\cellcolor{red!15}\textbf{-12.54}\\
        & $\text{Llama-3-70b}$~\cite{meta2024} & 58.39 &\cellcolor{red!15}\textbf{-9.42} \\
        & $\text{GPT-4-0613}$~\cite{gpt4} & 60.10 &\cellcolor{red!15}\textbf{-13.97} \\
        & $\text{Gemini-1.5-Pro}$~\cite{gemini}  &62.85 &\cellcolor{red!15}\textbf{-9.11} \\
        & $\text{Mixtral-8x22B}$~\cite{mixtral} &65.11 &\cellcolor{green!5}+0.57 \\
        \midrule
        \multirow{2}{*}{$\text{ImageLLM}^*$}  & Gemini-1.5-Pro~\cite{gemini} &58.29 &\cellcolor{red!15}\textbf{-10.68}  \\
        & GPT-4o~\cite{gpt4} &60.41 &\cellcolor{red!15}\textbf{-9.92} \\
        \midrule
        \multirow{4}{*}{$\text{VideoLLM}^*$} 
        & Minigpt-4~\cite{videochat} &61.69  &\cellcolor{green!5}-1.02 \\
        & VideoChat2~\cite{mvbench} &65.28  &\textbf{-5.35}\\
        & PLLaVA~\cite{pllava} &67.01 &\cellcolor{green!5}-2.13\\
        & VideoLLaVA~\cite{videoLLaVA} &67.15 &\cellcolor{green!5}+2.50  \\

        \midrule
        \multirow{2}{*}{$\text{Base}^{\dag} $}      
        &$\text{Videobert}$~\cite{videobert} &60.74 &\cellcolor{green!5}-1.59\\
        &\text{CLIP (ViT-L/14)}~\cite{clip} &63.00  &\cellcolor{green!5}-0.88  \\
        \midrule
        $\textbf{Ours}^{\dag}$
        & \textbf{VGCM} &\textbf{\underline{76.23}}  &\cellcolor{green!5}-0.35 \\
        \bottomrule
      \end{tabular}
    }}
    \label{tab:illusory_test}
\end{table}
\begin{figure}[t]
\begin{center}
\includegraphics[width=0.48\textwidth]{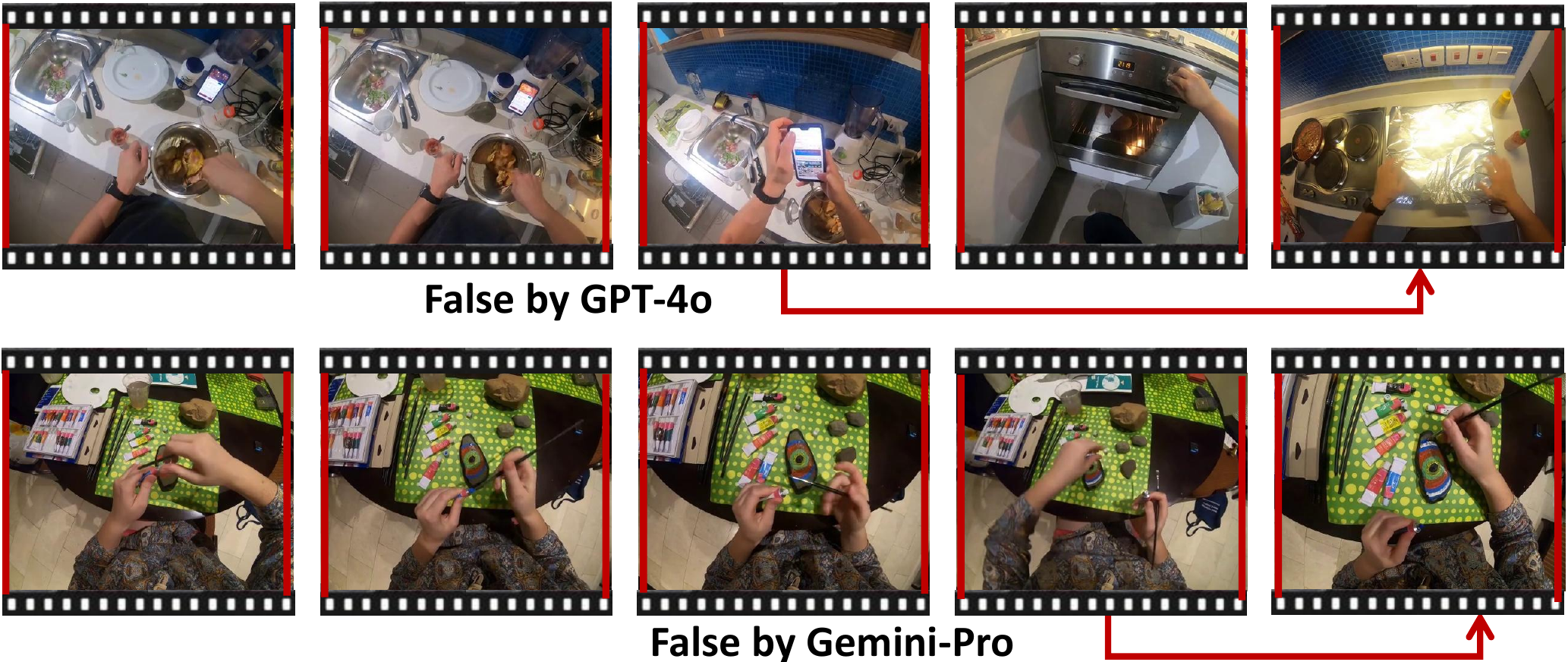}
\end{center}
\caption{\textbf{Illusory Test Visualization Examples.} LLMs and ImageLLMs suffer from serious illusory problems, leading to erroneous reasoning of non-existent causal relations between similar objects and scenes.}
\label{fig:illu}
\end{figure}

\noindent\textbf{e) Illusory Test}
To further assess the hallucination tendencies, we collect an additional validation set of 100 examples where causality does not exist between any events. 
Some examples include pairs of events with frequent statistical regularity, while others include cases where the prior event only serves as a necessary condition but does not directly lead.

As shown in Tab.~\ref{tab:illusory_test}, we can observe that the performances of LLMs drop significantly when judging the causal relations between examples with conditional correlation or statistical correlation in the temporal domain.
Visualization examples are shown in Fig.~\ref{fig:illu}, in example one, GPT-4o mistakenly believes that several unrelated cooking steps are causal. 
Although OpenAI~\cite{gpt4} mentioned that GPT’s upgrading process reduces the hallucination issue in various other tasks, both GPT-4 and GPT-4o still show serious hallucination problems. 
In comparison, VideoLLMs and VGCM exhibit reduced susceptibility to illusions with the introduction of extra visual information. 

\subsection{Ablation Study}
We conduct ablation studies to evaluate various components of the VGCM model, including the loss function, encoder design, causal methods, and input modalities, as described in Sections \textbf{a) to d)}. 
Besides we perform a detailed analysis of the architecture and input modalities of VideoLLMs. 
Additionally, we examine the annotation subjectivity and data volume of the MECD dataset.

\begin{table}[t]
    \centering
        \caption {\textbf{Ablation Study.} Adj indicates Front-door adjustment, Inter indicates Counterfactual intervention, and Context indicates context chain reasoning. }
    \resizebox{0.48\textwidth}{!}{
      \setlength{\tabcolsep}{1.5mm}
  \begin{tabular}{cccccccccc}
    \toprule
     \multicolumn{3}{c}{$\textbf{Base designs}$} & \multicolumn{3}{c}{$\textbf{Causal methods}$} 
     & \multicolumn{2}{c}{$\textbf{Metircs}$}  \\
     $\mathcal{L}_{C}$ & $\mathcal{L}_{V}$ & $\mathcal{L}_{S}$ & Adj & Inter & Context &Acc~$\uparrow$ &SHD~$\downarrow$\\
    \midrule
     &\checkmark &\checkmark  & & & & 64.8 &4.66\\
    \checkmark &  &\checkmark  & & & & 65.1 &4.71\\
    \checkmark & \checkmark &  & & & & 65.3 &4.64\\
    \checkmark & \checkmark & \checkmark & & & & 67.0 &4.59\\
    \checkmark & \checkmark & \checkmark & \checkmark & & & 68.7 &4.27\\
    \checkmark & \checkmark & \checkmark & & \checkmark & & 69.3 &4.32\\
    \checkmark & \checkmark & \checkmark & \checkmark & \checkmark &  & 71.2 &4.19\\
    \checkmark & \checkmark & \checkmark & \checkmark & \checkmark & \checkmark & 71.3 &3.94\\
    \bottomrule
  \end{tabular}
  \label{tab:ablationresults}
}
\end{table}
\noindent\textbf{a) Loss Function.}
We design our causal discovery model based on Granger Causality and applied three auxiliary losses $\mathcal{L}_V$, $\mathcal{L}_C$, and $\mathcal{L}_S$ to enhance its reasoning capabilities. 
As shown in Tab.~\ref{tab:ablationresults}, the benefits of $\mathcal{L}_V$ and $\mathcal{L}_C$ on our VGCM model are evident, as they facilitate event prediction, which in turn supports the inference of causal relations. 
A stronger event prediction ability enables the model to better determine whether the existence of a certain event is beneficial for predicting the result event, thereby inferring causal relations. 
Additionally, $\mathcal{L}_S$ contributes by supervising the prediction of $e_N$ with and without certain non-causal event $e_k$ masked, ensuring that the prediction of the result event is independent of the occurrence or non-occurrence of causally unrelated premise events.

\begin{table}[t]
    \centering
        \caption {\textbf{Encoder Ablation Study.} Videobert is utilized as the encoder by default, we also compared the performance of utilizing other multi-modal pre-trained encoders such as CLIP.}
    \resizebox{0.48\textwidth}{!}{
      \setlength{\tabcolsep}{3.0mm}{
      \begin{tabular}{llcc}
        \toprule
        &Encoder &SHD~$\downarrow$ &Acc~$\uparrow$ \\
        \midrule
        \multirow{5}{*}{\textbf{VGCM}}
        & \textbf{Univl}~\cite{univl} &4.24  &68.33  \\
        & \textbf{CLIP-B}~\cite{clip} &4.15 &69.10  \\
        & \textbf{CLIP-L}~\cite{clip} &4.17 &69.99  \\
        & \textbf{All-in-one-B+}~\cite{all} &4.24  &70.01  \\
        & \textbf{Videobert}~\cite{videobert} &\textbf{3.94} & \textbf{71.28} \\
        \bottomrule
      \end{tabular}
        \label{tab:ablationencoder}
    }}
\end{table}
\noindent\textbf{b) Encoder.}
For our encoders, $\text{Enc}_V$ and $\text{Enc}_C$, we explored the possibility of substituting them with other vision-language pre-trained models, including Univl~\cite{univl}, CLIP~\cite{clip}, and All-in-one-B+~\cite{all} as indicated in Tab.~\ref{tab:ablationencoder}. 

The highest performance was achieved when Videobert~\cite{videobert} was employed as the encoder, the core reason is as follows. 
{The mask-prediction training approach in Videobert~\cite{videobert} is particularly effective for learning contextual relations, making it well-suited for subsequent causal reasoning guided by Granger Causality. The principle of Granger Causality is to assess whether preceding event information improves the prediction of future outcomes—a concept that aligns with Videobert’s objective of modeling contextual dependencies.}


\noindent\textbf{c) Causal Methods.}

\begin{table}[t]
    \centering
        \caption {\textbf{Illusory Temporal Causality Experiment.} w/o F indicates without Front-door Adjustment, the minor accuracy drops of particular position indicate the illusory is released.}
    \resizebox{0.48\textwidth}{!}{
      \setlength{\tabcolsep}{1.3mm}{
       \begin{tabular}{lccc}
        \toprule
        Method &$\Delta$Acc of $r_0$& $\Delta$Acc of $ r_{N-1} $&Overall Acc\\
        \midrule
        VAR~\cite{VAR} &\cellcolor{red!15}-3.5 &\cellcolor{red!15}-3.7 &57.3\\
        $\text{VGCM (w/o F)}$ &\cellcolor{red!15}-3.3 &\cellcolor{red!15}-3.2 &66.9\\
    \textbf{$\text{VGCM}$} & \cellcolor{green!5}\textbf{\underline{-0.7}} &\cellcolor{green!5}\textbf{\underline{-0.3}} &68.7\\
        \bottomrule
      \end{tabular}
        \label{tab:temporal_experiment}
    }}
\end{table}

\begin{table}[t]
    \centering
        \caption {\textbf{Illusory Existence Causality Experiment.} w/o C indicates without counterfactual intervention, the similarity division results indicate that illusory is suppressed as models pay more attention to causality rather than simple semantics.}
    \resizebox{0.48\textwidth}{!}{
      \setlength{\tabcolsep}{2.0mm}{
       \begin{tabular}{lcc}
        \toprule
        Method & starting division & Ending division\\
        \midrule
        $\text{VGCM (w/o C)}$ &1.12 &1.04\\
    \textbf{$\text{VGCM}$} &1.12 & \textbf{\underline{0.93}} \\
        \bottomrule
      \end{tabular}
        \label{tab:distance}
    }}
\end{table}

\noindent\textbf{1) Front-door Adjustment.}
The method does improve reasoning ability in Tab.~\ref{tab:ablationresults}. We conduct an experiment in Tab.~\ref{tab:temporal_experiment} for further proof. Since events closer to the result event are higher as the cause, the model likely learns these biased time-domain tendencies.
So we compare the accuracy of VGCM without front-door adjustment and VGCM in determining the first relation $r_1$ and the last relation $r_{N-1}$.
The results demonstrate that temporal illusory causality is greatly mitigated. 

\noindent\textbf{2) Counterfactual Intervention.}
The performance in Tab.~\ref{tab:ablationresults} shows that counterfactual intervention with existence-only descriptions does facilitate the model with powerful reasoning ability.
We dive into further analysis on the basis that when a non-causal event is masked, the causal feature ${\boldsymbol F}^m_k$ fed into the causal relation head should be similar to the unmasked feature ${\boldsymbol F}^p$, instead, a bigger gap appears when masking a causal event. For stronger proof, we measure the difference in feature similarity in Tab.~\ref{tab:distance}. 
We define the similarities division as the quotient of the similarity(${\boldsymbol F}^m_k$, ${\boldsymbol F}^p$) with a non-causal $e_k$ masked over with a causal $e_k$ masked. In the experiment, we find that the similarity division is always above 1 without the counterfactual intervention, however, the division is below 1 with the help of counterfactual intervention.

\noindent\textbf{3) Context Chain Reasoning.}
The results presented in Tab.~\ref{tab:ablationresults} indicate that, despite not enhancing the accuracy of causal chain discovery, context chain reasoning enhances the model's overall causal reasoning ability obviously.

\noindent\textbf{d) Input Modalities.}
Typically, the input of VGCM consists of a textual input with an average of 13.5 words and a visual input of 50 frames. To investigate the influence of the text modality, we employ a masking strategy for the input caption of the premise event, gradually increasing the masking ratio from 10\% to 80\%. 
The results presented in Fig.~\ref{mask_fig} indicate that VGCM does not rely on the textual modality input.

\begin{figure}[t]
    \centering
        \caption {\textbf{Experiment of Input Modalities of the VGCM.} The x-axis represents the proportion of masked textual input, and the y-axis represents the accuracy of causal reasoning. When 80\% of textual or visual information is masked, VGCM still performs well in inferring causal relations, and visual information plays a more significant role.}
\begin{tikzpicture}
\begin{axis}[
    x tick label style={/pgf/number format/1000 sep=},
    scale=0.65,
    symbolic x coords={0.1, 0.3, 0.4, 0.8, 1},
    enlargelimits=0.05,
    legend style={at={(0.5,-0.1)}, anchor=north, legend columns=-1, yshift=-8pt},
    ybar interval=0.7,
    xmin=0.1, xmax=1,
    ymax=72,  
    ymin=66,  
]
\draw[dashed, color=red, line width=1.2pt] (axis cs:0.1,71.2) -- (axis cs:0.8,71.2);

\addplot 
	coordinates {(0.1,71.2) (0.3,71.2)
		 (0.4,71.2) (0.8,71.2) (1,71.2)};
\addplot 
	coordinates {(0.1,70.7) (0.3,69.9)
		 (0.4,69.5) (0.8,69.0) (1,71.2)};
\addplot 
	coordinates {(0.1,70.3) (0.3,69.0)
		 (0.4,68.3) (0.8,67.9) (1,71.2)};
\legend{Non-masked, Text-masked, Vision-masked}
\end{axis}
\end{tikzpicture}
\label{mask_fig}
\end{figure}

In contrast, the experimental results suggest a more obvious performance decrease towards less visual modality input in the causality discovery task, as shown in Fig.~\ref{mask_fig}. 
\begin{table}[t]
    \centering
        \caption {\textbf{Performance Comparison between the Base Multi-modal Model and Corresponding VideoLLM.} An obvious gain is achieved after utilizing an LLM reasoner.}
    \resizebox{0.48\textwidth}{!}{
      \setlength{\tabcolsep}{3.7mm}{
      \begin{tabular}{lcccl}
        \toprule
         Method &SHD~$\downarrow$ &Accuracy~$\uparrow$ \\
        \midrule
        \text{EVA-CLIP}~\cite{evaclip} &4.75 &65.76   \\
         VideoChat2~\cite{mvbench} &4.68 &\cellcolor{green!5}68.58~\textbf{(+2.82)} \\
        \midrule
        
        \text{CLIP}~\cite{clip} &4.77 &62.87  \\
         PLLaVA~\cite{pllava} &4.74  &\cellcolor{green!5}65.71~\textbf{(+2.84)} \\
        \midrule
        
        LanguageBind~\cite{languagebind} &4.92  &59.83 \\
         VideoLLaVA~\cite{videoLLaVA} &4.73 &\cellcolor{green!5}67.12~\textbf{(+7.29)} \\
        \bottomrule
      \end{tabular}
    }}
    \label{multi-vllm}
\end{table}

\begin{table}[t]
    \centering
        \caption {\textbf{Performance Comparison between VideoLLM with and without Textual Input.} Minor increments indicate a high Visual-Textual Information Ratio of the MECD dataset. }
    \resizebox{0.48\textwidth}{!}{
      \setlength{\tabcolsep}{0.8mm}{
      \begin{tabular}{llccl}
        \toprule
          &Method &SHD~$\downarrow$ &Pos~$\uparrow$ &Accuracy~$\uparrow$ \\
        \midrule
        \multirow{5}{*}{V} 
        & MiniGPT4-Video~\cite{minigpt4} &5.15 &54.90  &57.45 \\
        & MiniGPT-4~\cite{videochat} &5.14 &52.89 & 57.75  \\
        & VideoLLaVA~\cite{videoLLaVA} &5.10 &60.44 & 60.58 \\
        & PLLaVA~\cite{pllava} &5.03 &57.10  &60.97 \\
        & VideoChat2~\cite{mvbench} &4.88 &57.42 &63.03\\
        \midrule
        \multirow{5}{*}{V + T} 
        & MiniGPT-4~\cite{videochat} &5.12&\cellcolor{red!15}51.92 & \cellcolor{green!5}58.29 \textbf{(+0.54)}\\
        & MiniGPT4-Video~\cite{minigpt4} &5.00 &\cellcolor{red!15}53.85 &\cellcolor{green!5}59.88 \textbf{(+2.43)} \\
        & VideoLLaVA~\cite{videoLLaVA} &4.95 &\cellcolor{red!15}57.46 & \cellcolor{green!5}62.78 \textbf{(+2.20)}\\
        & PLLaVA~\cite{pllava} &4.86 &\cellcolor{red!15}55.88 &\cellcolor{green!5}63.00 \textbf{(+2.03)}\\
        & VideoChat2~\cite{mvbench} &4.75 &\cellcolor{red!15}55.57 &\cellcolor{green!5}63.85 \textbf{(+0.82)}\\
        \bottomrule
      \end{tabular}
    }}
    \label{only-v}
    \vspace{-5pt}
\end{table}
\noindent\textbf{e) Architecture and Input Analysis of VideoLLMs.} We analyze the impact of the LLM module and input modalities in VideoLLMs, focusing on two aspects: how LLMs enhance reasoning abilities compared to simple MLP layers, and the visual-to-textual information ratio in the MECD dataset.

\noindent\textbf{1) Reasoning Abilities with LLMs Versus MLPs.}
We compare the performances between VideoLLMs and corresponding basic multi-modal models in Tab.~\ref{multi-vllm}. 
We found the understanding capabilities of LLMs significantly enhance the causal discovery of multi-modal foundation models versus simple MLP layers.
Although visual and textual information are encoded by the same encoder, the models are equipped with enhanced reasoning capabilities through a more sophisticated 7B language model. 

\begin{figure}[t]
\begin{center}
\includegraphics[width=0.48\textwidth]{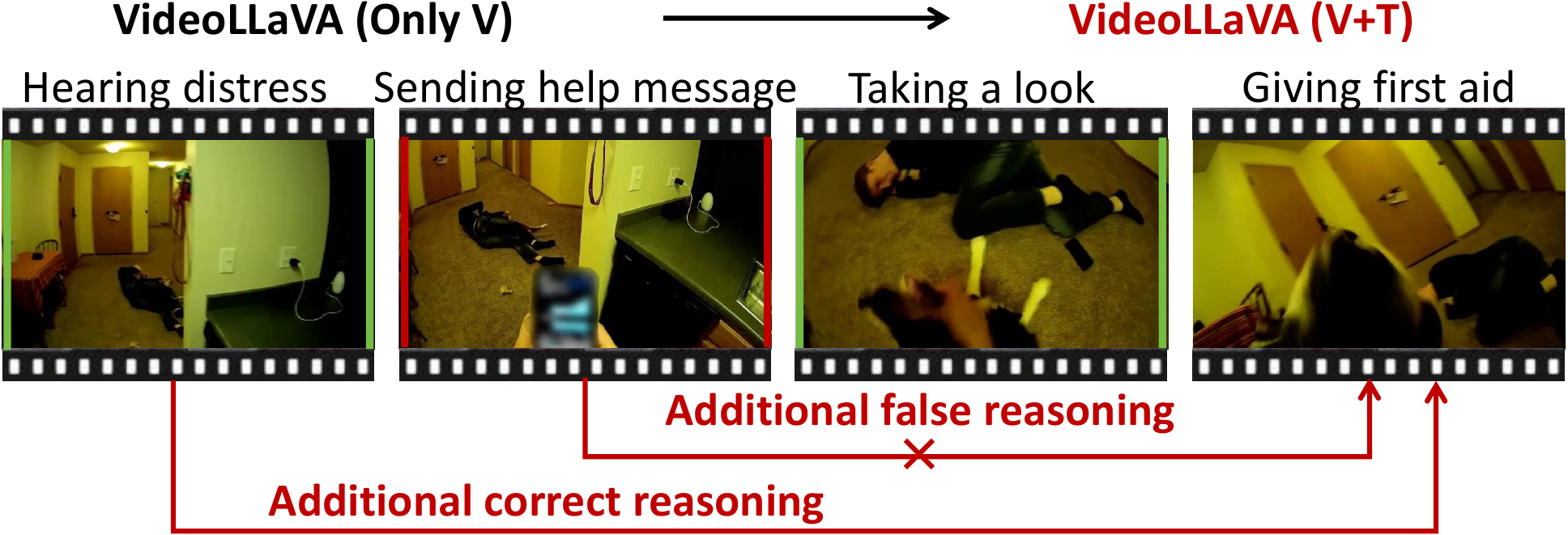}
\end{center}
\caption{{\textbf{Input Modality Test of VLLMs.}} When prompting VideoLLaVA with both textual and visual modalities compared with only visual input, VideoLLaVA successfully discovered more causal relations, however illusory problem arises.}
\label{fig:vt}
\vspace{-5pt}
\end{figure}

\noindent\textbf{2) Input Modalities of VideoLLMs.} We compare the performances between VideoLLMs with all modalities input and only vision input as shown in Tab.~\ref{only-v}. 
When additional text information is input into VideoLLMs, it serves to supplement the limitations of visual information and emphasizes key details. 
However, ambiguous text cues can exacerbate illusions, which in turn leads to a decline in the Pos metric that contrasts with the Acc trend. 
The modest increase in the overall indicator of causal reasoning ability (+1.43) underscores a high Visual-Textual Information Ratio of the MECD dataset~\cite{mvbench}. 

As shown in Fig.~\ref{fig:vt}, when prompted by additional caption information, although a missing causal relation is discovered, the illusory causality problem of VideoLLaVA~\cite{videoLLaVA} becomes more serious. 
A more serious hallucination problem led to the mistaken perception that ``Sending help messages'' would result in ``Giving first aid''.

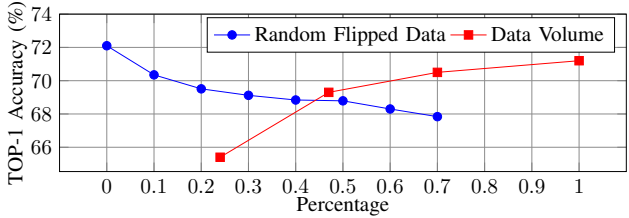
\begin{figure}[t]
\begin{tikzpicture}[scale=0.8]
\begin{axis}[
    xlabel={Percentage},
    ylabel={TOP-1 Accuracy (\%)},
    xlabel style={yshift=5pt}, 
    ylabel style={yshift=-5pt}, 
    xtick={0,0.1,0.2,0.3,0.4,0.5,0.6,0.7,0.8,0.9,1.0},
    ytick={60,62,64,66,68,70,72,74},
    ymax = 74,
    legend style={at={(0.98,0.98)}, anchor=north east, legend columns=2},
    grid=both,
    grid style={line width=0.5pt, draw=gray!50},
    major grid style={line width=0.3pt,draw=gray!90},
    width=11.0cm,
    height=4.2cm
]

\addplot[color=blue,mark=*] coordinates {
    (0.0, 72.10)
    (0.1, 70.35)
    (0.2, 69.51)
    (0.3, 69.12)
    (0.4, 68.84)
    (0.5, 68.79)
    (0.6, 68.30)
    (0.7, 67.84)
};

\addplot[color=red,mark=square*] coordinates {
    (0.24, 65.4)
    (0.47, 69.3)
    (0.7, 70.5)
    (1.0, 71.2)
};

\legend{
    Random Flipped Data,
    Data Volume
}
\end{axis}
\end{tikzpicture}
\vspace{-5pt}
\caption{\textbf{Dataset Robustness.} Accuracy decreases slightly when increasing noise to the causal relation annotations of the training set, and increases slowly when training data increases.}
\label{quant}
\vspace{-5pt}
\end{figure}

\noindent\textbf{f) Dataset Annotation and Volume.} 
We study the subjectivity and data volume of our proposed MECD dataset, which is shown in Fig.~\ref{quant}. 
In the experiments of increasing the ratio of randomly flipped annotated causal relations (flipping only one relation of the whole causal relations of video), the accuracy decreases slightly, demonstrating the small amount of subjectivity in labeling does not have a serious impact. 
Besides, we analyze the scale of training data, the increment from 1,000 examples to 1,139 examples yields a very modest improvement, indicating the adequacy of MECD. 

\subsection{Downstream Tasks}
Moreover, to further validate the generalized reasoning capabilities of our model, we evaluate the quality of output causal relations on two related and representative video reasoning tasks: Video Question Answering (VQA) and Event Prediction (EP) as shown in Tab.~\ref{tab:downstream_vqa} and Tab.~\ref{tab:downstream_ep}.

\noindent\textbf{a) Downstream Task-VQA.} 
We identified 527 examples from the ActivityNet VQA dataset ~\cite{activitynetqa} that overlap with the MECD test set, which we used to evaluate the VQA task for this subset that has been separated. 
Specifically, we prompted MiniGPT4-video~\cite{minigpt4} with additional causal relations outputs alongside the standard question inputs. This paradigm facilitates the VideoLLMs in considering the task from a causal perspective. 
As shown in the Tab.~\ref{tab:downstream_vqa}, when prompted with these additional causal relations, the answering accuracy of MiniGPT4-video~\cite{minigpt4} improved by VGCM surpasses VLLMs. 
These findings confirm that VGCM can provide accurate causal perception for videos, significantly improving performance on related video reasoning tasks concerning explanation, description, and prediction~\cite{CLEVRER}.

\begin{table}[t]
    \centering
        \caption {\textbf{Downstream Video Question Answering Task.} Higher VQA Acc and VQA Score are reached when prompted with causal relations reasoned from our VGCM.}
    \resizebox{0.48\textwidth}{!}{
      \setlength{\tabcolsep}{1.3mm}{
      \begin{tabular}{lcc}
        \toprule
        Output Causal Relations  & VQA Acc & VQA Score \\
        \midrule
        w/o (Standard QA setting for VLLMs) & 43.17 &  2.82 \\
        w Gemini-Pro~\cite{gemini} & 49.10  &  2.90  \\
        w GPT-4~\cite{gpt4}  & 49.36 &  2.89  \\
        w VideoChat2~\cite{mvbench}~(Vicuna-7B) & 51.01 &\textbf{2.97} \\
        w VideoLLaVA~\cite{videoLLaVA}~(Vicuna-7B) & 51.88 & 2.93 \\
        w PLLaVA~\cite{pllava}~(Vicuna-7B) & 52.47 &2.96 \\
        w VideoChat2~\cite{mvbench}~(Mistral-7B) &\textbf{55.42} & 2.93 \\
        \textbf{\underline{w Our VGCM}} & \textbf{\underline{62.21}} & \textbf{\underline{3.12}} \\
        \bottomrule
      \end{tabular}
    }}
    \label{tab:downstream_vqa}
      \end{table}
      
\begin{figure}[t]
\begin{center}
\includegraphics[width=0.48\textwidth]{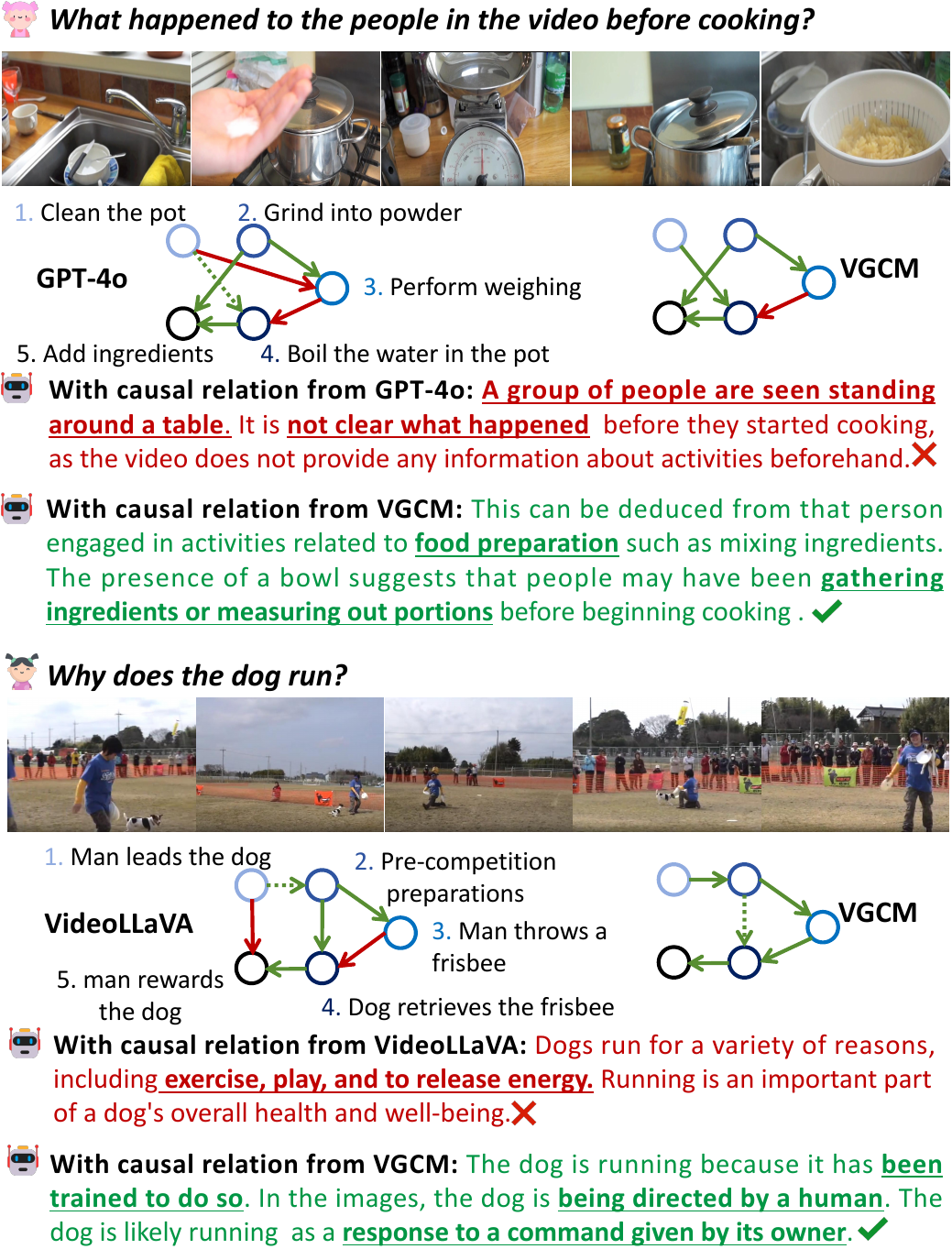}
\end{center}
    \caption{\textbf{{Visualizations for Downstream Video Question Answering.}} VideoLLMs offer more reasonable answers with the additional help of causal graphs reasoned from VGCM.}
    \label{vis_vqa}
    \vspace{-5pt}
\end{figure}

The visualization of the QA task's comparison results, as depicted in Fig.~\ref{vis_vqa}, illustrates the first example as a temporal order question. 
Prompted with the causal graph generated by our model, the model correctly infers that food preparation should precede cooking. 
Conversely, when prompted with the causal graph produced by VideoLLaVA~\cite{videoLLaVA}, the answers and temporal relations are chaotic.

\noindent\textbf{b) Downstream Task-Event Prediction}.
\begin{table}[t]
    \centering
        \caption {\textbf{Downstream Event Prediction Task.} Higher Event prediction Acc and Event prediction Score are reached when prompted with causal relations from our VGCM.}
    \resizebox{0.48\textwidth}{!}{
      \setlength{\tabcolsep}{1.3mm}{
      \begin{tabular}{lcc}
        \toprule
        Output Causal Relations  & EP Acc & EP Score \\
        \midrule
        w/o (Standard EP setting for VLLMs) &20.83   & 2.08 \\
        w Gemini-Pro~\cite{gemini} &28.44  &2.54 \\
        w GPT-4~\cite{gpt4}  &29.32  &2.60  \\
        w VideoChat2~\cite{mvbench}~(Vicuna-7B) &28.66 &2.63 \\
        w VideoLLaVA~\cite{videoLLaVA}~(Vicuna-7B) & 28.71 & 2.61 \\
        w PLLaVA~\cite{pllava}~(Vicuna-7B) &36.98 & 2.67\\
        w VideoChat2~\cite{mvbench}~(Mistral-7B) &\textbf{39.39} & \textbf{2.71} \\
        \textbf{\underline{w Our VGCM}} & \textbf{\underline{43.39}} & \textbf{\underline{2.73}} \\
        \bottomrule
      \end{tabular}
    }}
    \label{tab:downstream_ep}
      \end{table}
Similar to the VQA task, we introduce a new task: Event Prediction. 
We utilize the first n-1 events from the MECD test set as known events, requiring the model to predict the content of the subsequent n-th event based on the acquired information. 
The evaluation method is the same as for the VQA task: we use the GPT-4 API to score the prediction results against the ground truth event and determine whether the prediction is correct. We prompted models with the following instruction \texttt{``Please answer that what could happen next with the highest probability?''}.

Specifically, we adopted a strategy similar to the VQA task, prompting MiniGPT4-video~\cite{minigpt4} with additional causal relation outputs in addition to the standard question inputs. 
As shown in Tab.~\ref{tab:downstream_ep}, when prompted with these additional causal relations, the event prediction accuracy of MiniGPT4-video~\cite{minigpt4}, enhanced by our VGCM, outperformed other strong VLLMs. 
These results confirm that our model can also enhance performance on related event prediction tasks. 

\begin{figure}[t]
\begin{center}
\includegraphics[width=0.48\textwidth]{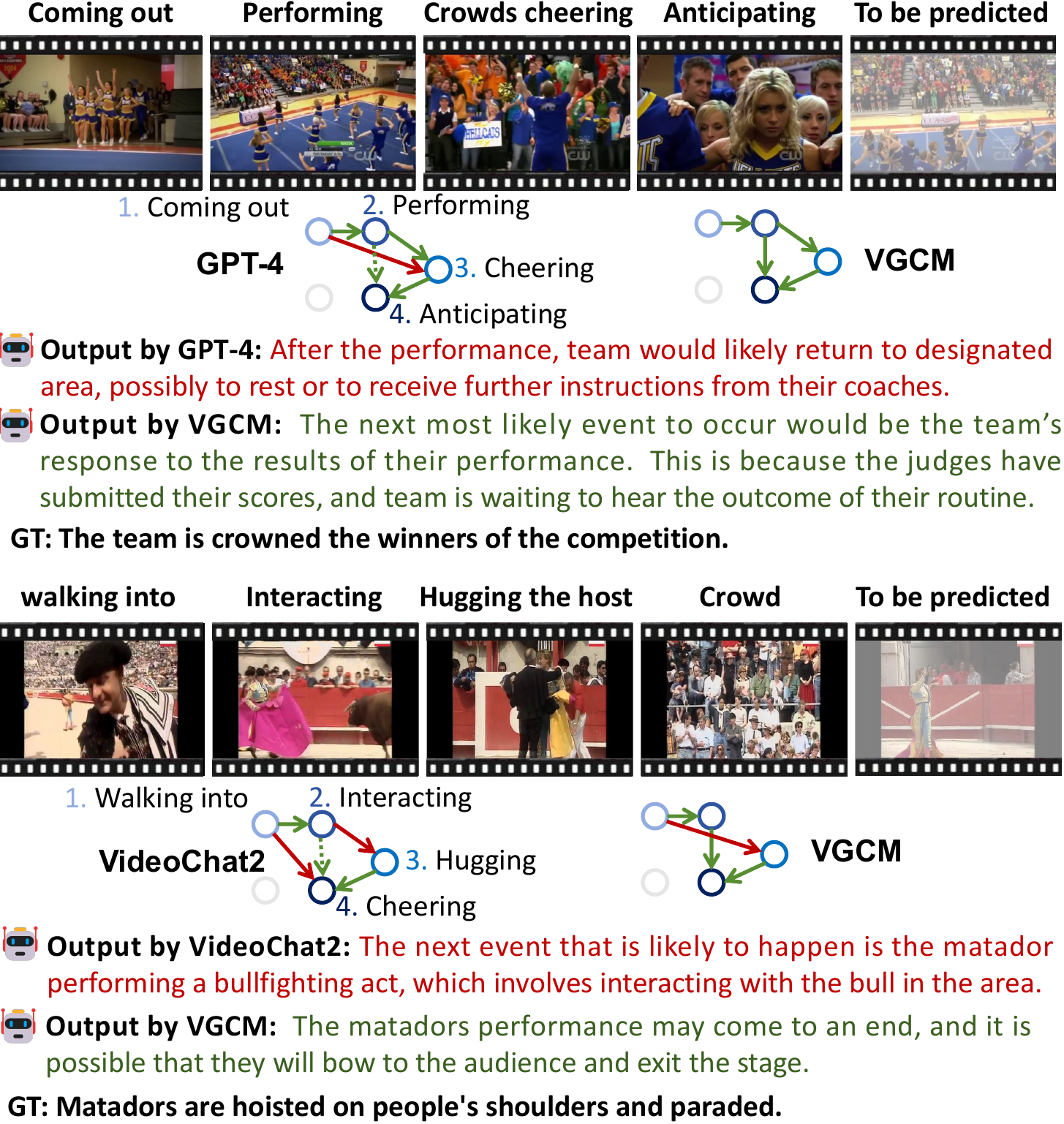}
\end{center}
    \caption{\textbf{{Visualizations for the Downstream Event Prediction Task.}} VideoLLMs offer more reasonable event prediction results with the help of causal graphs output from our VGCM.}
    \label{vis_ep}
\end{figure}

As shown in Fig.~\ref{vis_ep}, when prompting with the causal graph output from our VGCM, the model accurately infers from a series of prior events that the players have delivered a strong performance. 
It likely interprets the players' nervous expressions as anticipation of the game results and their reactions to those results.
In contrast, with the causal graph generated by GPT-4~\cite{gpt4}, the model makes incorrect predictions based on common assumptions, such as the idea that the team would receive further instructions from their coaches.

\section{Conclusion}
We proposed a novel task, multi-event video causal discovery (MECD), which focuses on event-level complete causal graph reasoning. 
To support this task, we developed the MECD dataset, consisting of 1,438 long-term daily-life videos featuring intricate causal relations. 
Additionally, we proposed the first model for video causal discovery, the Video Granger Causality Model (VGCM), which is based on the principles of the \textit{Event Granger Test} and incorporates advanced causal inference techniques to address issues such as illusions and confounding factors. 
VGCM employs robust and efficient reasoning mechanisms, including context chain reasoning and non-regressive complete graph reasoning. 
Our VGCM outperformed GPT-4o and VideoChat2 by 5.77\% and 2.70\%, respectively, highlighting its superior reasoning capabilities.

\bibliographystyle{IEEEtran}
\bibliography{main}

\begin{thebibliography}{100}
\providecommand{\url}[1]{#1}
\csname url@samestyle\endcsname
\providecommand{\newblock}{\relax}
\providecommand{\bibinfo}[2]{#2}
\providecommand{\BIBentrySTDinterwordspacing}{\spaceskip=0pt\relax}
\providecommand{\BIBentryALTinterwordstretchfactor}{4}
\providecommand{\BIBentryALTinterwordspacing}{\spaceskip=\fontdimen2\font plus
\BIBentryALTinterwordstretchfactor\fontdimen3\font minus \fontdimen4\font\relax}
\providecommand{\BIBforeignlanguage}[2]{{%
\expandafter\ifx\csname l@#1\endcsname\relax
\typeout{** WARNING: IEEEtran.bst: No hyphenation pattern has been}%
\typeout{** loaded for the language `#1'. Using the pattern for}%
\typeout{** the default language instead.}%
\else
\language=\csname l@#1\endcsname
\fi
#2}}
\providecommand{\BIBdecl}{\relax}
\BIBdecl

\bibitem{driving}
S.~Park, M.~Lee, and e.~a. Kang, JiHyuk, ``Vlaad: Vision and language assistant for autonomous driving,'' in \emph{WACV}, 2024, pp. 980--987.

\bibitem{act}
J.~Zhang, F.~Shen, X.~Xu, and H.~T. Shen, ``Temporal reasoning graph for activity recognition,'' \emph{TIP}, vol.~29, pp. 5491--5506, 2020.

\bibitem{automatic}
N.~M. Robertson and I.~D. Reid, ``Automatic reasoning about causal events in surveillance video,'' \emph{EURASIP Journal on Image and Video Processing}, vol. 2011, pp. 1--19, 2011.

\bibitem{robot}
Q.~Liu, G.~Wang, Z.~Liu, and H.~Wang, ``Visuomotor navigation for embodied robots with spatial memory and semantic reasoning cognition,'' \emph{TNNLS}, 2024.

\bibitem{SeViLA}
S.~Yu, J.~Cho, and e.~a. Yadav, Prateek, ``Self-chained image-language model for video localization and question answering,'' \emph{arXiv preprint arXiv:2305.06988}, 2023.

\bibitem{NEXT}
J.~Xiao, X.~Shang, A.~Yao, and T.-S. Chua, ``Next-qa: Next phase of question-answering to explaining temporal actions,'' in \emph{CVPR}, 2021, pp. 9777--9786.

\bibitem{CLEVRER}
K.~Yi, C.~Gan, and e.~a. Li, ``Clevrer: Collision events for video representation and reasoning,'' \emph{arXiv preprint arXiv:1910.01442}, 2019.

\bibitem{nmulti1}
R.~Choudhury, K.~Niinuma, and e.~a. Kitani, Kris~M, ``Video question answering with procedural programs,'' in \emph{ECCV}.\hskip 1em plus 0.5em minus 0.4em\relax Springer, 2025, pp. 315--332.

\bibitem{nmulti2}
A.~Yang, A.~Miech, and e.~a. Sivic, Josef, ``Just ask: Learning to answer questions from millions of narrated videos,'' in \emph{ICCV}, 2021, pp. 1686--1697.

\bibitem{nmulti3}
R.~Choudhury, K.~Niinuma, K.~M. Kitani, and L.~A. Jeni, ``Zero-shot video question answering with procedural programs,'' \emph{arXiv preprint arXiv:2312.00937}, 2023.

\bibitem{nextgqa}
J.~Xiao and e.~a. Yao, Angela, ``Can i trust your answer? visually grounded video question answering,'' in \emph{CVPR}, 2024, pp. 13\,204--13\,214.

\bibitem{videotree}
Z.~Wang, S.~Yu, and e.~a. Stengel-Eskin, ``Videotree: Adaptive tree-based video representation for llm reasoning on long videos,'' \emph{arXiv preprint arXiv:2405.19209}, 2024.

\bibitem{timecraft}
L.~Huabin, M.~Xiao, Z.~Cheng, Z.~Yang, and L.~Weiyao, ``Timecraft: Navigate weakly-supervised temporal grounded video question answering via bi-directional reasoning,'' in \emph{ECCV}, 2024.

\bibitem{BiGED}
C.~Tan, C.~K. Yeo, C.~Tan, and B.~Fernando, ``Abductive action inference,'' \emph{arXiv preprint arXiv:2210.13984}, 2022.

\bibitem{VAR}
C.~Liang, W.~Wang, T.~Zhou, and Y.~Yang, ``Visual abductive reasoning,'' in \emph{CVPR}, 2022, pp. 15\,565--15\,575.

\bibitem{rextime}
J.-J. Chen, Y.-C. Liao, and e.~a. Lin, Hsi-Che, ``Rextime: A benchmark suite for reasoning-across-time in videos,'' \emph{arXiv preprint arXiv:2406.19392}, 2024.

\bibitem{granger}
A.~Seth, ``Granger causality,'' \emph{Scholarpedia}, vol.~2, no.~7, p. 1667, 2007.

\bibitem{granger2}
M.~Maziarz, ``A review of the granger-causality fallacy,'' \emph{The journal of philosophical economics: Reflections on economic and social issues}, vol.~8, no.~2, pp. 86--105, 2015.

\bibitem{granger3}
A.~Shojaie and E.~B. Fox, ``Granger causality: A review and recent advances,'' \emph{Annual Review of Statistics and Its Application}, vol.~9, pp. 289--319, 2022.

\bibitem{causalinference}
J.~Pearl, ``Causal inference,'' \emph{Causality: objectives and assessment}, pp. 39--58, 2010.

\bibitem{hanwangzhang}
X.~Yang, H.~Zhang, G.~Qi, and J.~Cai, ``Causal attention for vision-language tasks,'' in \emph{CVPR}, 2021, pp. 9847--9857.

\bibitem{count2}
W.~Wang, F.~Feng, and e.~a. He, Xiangnan, ``Clicks can be cheating: Counterfactual recommendation for mitigating clickbait issue,'' in \emph{SIGIR}, 2021, pp. 1288--1297.

\bibitem{cot1}
J.~Wei, X.~Wang, and e.~a. Schuurmans, Dale, ``Chain-of-thought prompting elicits reasoning in large language models,'' \emph{NeurIPS}, vol.~35, pp. 24\,824--24\,837, 2022.

\bibitem{cot2}
T.~Kojima, S.~S. Gu, and e.~a. Reid, Machel, ``Large language models are zero-shot reasoners,'' \emph{NeurIPS}, vol.~35, pp. 22\,199--22\,213, 2022.

\bibitem{cotsurvey}
Z.~Chu, J.~Chen, and e.~a. Chen, Qianglong, ``A survey of chain of thought reasoning: Advances, frontiers and future,'' \emph{arXiv preprint arXiv:2309.15402}, 2023.

\bibitem{MECD}
T.~Chen and e.~a. Liu, Huabin, ``{MECD}: Unlocking multi-event causal discovery in video reasoning,'' in \emph{NeurIPS}, 2024.

\bibitem{momentor}
L.~Qian, J.~Li, and e.~a. Wu, Yu, ``Momentor: Advancing video large language model with fine-grained temporal reasoning,'' \emph{arXiv preprint arXiv:2402.11435}, 2024.

\bibitem{videocot}
Y.~Wang and e.~a. Zeng, Yawen, ``{V}ideo{C}o{T}: A video chain-of-thought dataset with active annotation tool,'' in \emph{ALVR}, 2024, pp. 92--101.

\bibitem{videoofthought}
e.~a. Hao~Fei, Shengqiong~Wu, ``Video-of-thought: Step-by-step video reasoning from perception to cognition,'' in \emph{ICML}, 2024.

\bibitem{VideoEspresso}
S.~Han and e.~a. Wei~Huang, ``Videoespresso: A large-scale chain-of-thought dataset for fine-grained video reasoning via core frame selection,'' 2024.

\bibitem{Following}
R.~Hong and J.~Lang2411.14794, ``Following clues, approaching the truth: Explainable micro-video rumor detection via chain-of-thought reasoning,'' in \emph{WWW}, 2025.

\bibitem{glance}
Z.~Bai, R.~Wang, and X.~Chen, ``Glance and focus: Memory prompting for multi-event video question answering,'' \emph{NeurIPS}, vol.~36, 2024.

\bibitem{CATER}
R.~Girdhar and D.~Ramanan, ``Cater: A diagnostic dataset for compositional actions and temporal reasoning,'' \emph{arXiv preprint arXiv:1910.04744}, 2019.

\bibitem{AAR}
T.~Zhuo, Z.~Cheng, and e.~a. Zhang, Peng, ``Explainable video action reasoning via prior knowledge and state transitions,'' in \emph{ACM MM}, 2019, pp. 521--529.

\bibitem{LMLN}
Y.~Jin, L.~Zhu, and Y.~Mu, ``Complex video action reasoning via learnable markov logic network,'' in \emph{CVPR}, 2022, pp. 3242--3251.

\bibitem{s-relation}
Y.-H.~H. Tsai and e.~a. Divvala, Santosh, ``Video relationship reasoning using gated spatio-temporal energy graph,'' in \emph{CVPR}, 2019, pp. 10\,424--10\,433.

\bibitem{LPCMCI}
A.~Gerhardus and J.~Runge, ``High-recall causal discovery for autocorrelated time series with latent confounders,'' \emph{NeurIPS}, vol.~33, pp. 12\,615--12\,625, 2020.

\bibitem{PcGCE}
C.~K. Assaad, E.~Devijver, and E.~Gaussier, ``Discovery of extended summary graphs in time series,'' in \emph{Uncertainty in Artificial Intelligence}.\hskip 1em plus 0.5em minus 0.4em\relax PMLR, 2022, pp. 96--106.

\bibitem{NTS}
X.~Sun, O.~Schulte, and e.~a. Liu, Guiliang, ``Nts-notears: Learning nonparametric dbns with prior knowledge,'' \emph{arXiv preprint arXiv:2109.04286}, 2021.

\bibitem{DYNOTEARS}
R.~Pamfil and e.~a. Sriwattanaworachai, ``Dynotears: Structure learning from time-series data,'' in \emph{AISTATS}.\hskip 1em plus 0.5em minus 0.4em\relax PMLR, 2020, pp. 1595--1605.

\bibitem{THP}
R.~Cai and e.~a. Wu, Siyu, ``Thp: Topological hawkes processes for learning granger causality on event sequences,'' \emph{arXiv preprint arXiv:2105.10884}, 2021.

\bibitem{GC-nsHP}
W.~Chen and e.~a. Chen, Jibin, ``Learning granger causality for non-stationary hawkes processes,'' \emph{Neurocomputing}, vol. 468, pp. 22--32, 2022.

\bibitem{TCR}
Q.~Ning, Z.~Feng, H.~Wu, and D.~Roth, ``Joint reasoning for temporal and causal relations,'' \emph{arXiv preprint arXiv:1906.04941}, 2019.

\bibitem{RFBFN}
Z.~Li and e.~a. Fu, Luoyi, ``Rfbfn: A relation-first blank filling network for joint relational triple extraction,'' in \emph{ACL}, 2022, pp. 10--20.

\bibitem{CauSeRL}
X.~Zuo, P.~Cao, and e.~a. Chen, Yubo, ``Improving event causality identification via self-supervised representation learning on external causal statement,'' in \emph{ACL-IJCNLP}, 2021, pp. 2162--2172.

\bibitem{GESI}
C.~Fan, D.~Liu, L.~Qin, Y.~Zhang, and R.~Xu, ``Towards event-level causal relation identification,'' in \emph{Proceedings of the 45th international ACM SIGIR conference on research and development in information retrieval}, 2022, pp. 1828--1833.

\bibitem{CAUSE}
W.~Zhang, T.~Panum, S.~Jha, P.~Chalasani, and D.~Page, ``{CAUSE}: Learning {G}ranger causality from event sequences using attribution methods,'' in \emph{PMLR}, vol. 119, 2020, pp. 11\,235--11\,245.

\bibitem{ActivityNet}
R.~Krishna, K.~Hata, and e.~a. Ren, Frederic, ``Dense-captioning events in videos,'' in \emph{ICCV}, 2017, pp. 706--715.

\bibitem{egoschema}
K.~Mangalam, R.~Akshulakov, and J.~Malik, ``Egoschema: A diagnostic benchmark for very long-form video language understanding,'' \emph{NeurIPS}, vol.~36, pp. 46\,212--46\,244, 2023.

\bibitem{eventbench}
Y.~Du, K.~Zhou, Y.~Huo, and e.~a. Li, Yifan, ``Towards event-oriented long video understanding,'' \emph{arXiv preprint arXiv:2406.14129}, 2024.

\bibitem{videollama2}
Z.~Cheng, S.~Leng, and e.~a. Zhang, Hang, ``Videollama 2: Advancing spatial-temporal modeling and audio understanding in video-llms,'' \emph{arXiv preprint arXiv:2406.07476}, 2024.

\bibitem{gemini}
G.~Team and e.~a. Anil, Rohan, ``Gemini: a family of highly capable multimodal models,'' \emph{arXiv preprint arXiv:2312.11805}, 2023.

\bibitem{gpt4}
J.~Achiam, S.~Adler, and e.~a. Agarwal, Sandhini, ``Gpt-4 technical report,'' \emph{arXiv preprint arXiv:2303.08774}, 2023.

\bibitem{cross2}
W.~Lin, H.~Liu, and e.~a. Liu, Shizhan, ``Hieve: A large-scale benchmark for human-centric video analysis in complex events,'' \emph{IJCV}, vol. 131, no.~11, pp. 2994--3018, 2023.

\bibitem{cross3}
R.~N{\'e}meth, D.~Sik, and F.~M{\'a}t{\'e}, ``Machine learning of concepts hard even for humans: The case of online depression forums,'' \emph{International Journal of Qualitative Methods}, vol.~19, p. 1609406920949338, 2020.

\bibitem{sup1}
D.~Lopez-Paz, K.~Muandet, and B.~Recht, ``The randomized causation coefficient.'' \emph{J. Mach. Learn. Res.}, vol.~16, pp. 2901--2907, 2015.

\bibitem{sup2}
J.-F. Ton, D.~Sejdinovic, and K.~Fukumizu, ``Meta learning for causal direction,'' in \emph{AAAI}, 2021, pp. 9897--9905.

\bibitem{sup3}
H.~Li, Q.~Xiao, and J.~Tian, ``Supervised whole dag causal discovery,'' \emph{arXiv preprint arXiv:2006.04697}, 2020.

\bibitem{dvc1}
J.~Lei, L.~Wang, and e.~a. Shen, Yelong, ``Mart: Memory-augmented recurrent transformer for coherent video paragraph captioning,'' \emph{arXiv preprint arXiv:2005.05402}, 2020.

\bibitem{dvc2}
T.~Wang, R.~Zhang, and e.~a. Lu, Zhichao, ``End-to-end dense video captioning with parallel decoding,'' in \emph{ICCV}, 2021, pp. 6847--6857.

\bibitem{dvc3}
Z.~Zhang, Y.~Shi, and e.~a. Yuan, Chunfeng, ``Object relational graph with teacher-recommended learning for video captioning,'' in \emph{CVPR}, 2020, pp. 13\,278--13\,288.

\bibitem{ica1}
A.~Hyvarinen and H.~Morioka, ``Nonlinear ica of temporally dependent stationary sources,'' in \emph{Artificial Intelligence and Statistics}.\hskip 1em plus 0.5em minus 0.4em\relax PMLR, 2017, pp. 460--469.

\bibitem{ica2}
------, ``Unsupervised feature extraction by time-contrastive learning and nonlinear ica,'' \emph{NeurIPS}, vol.~29, 2016.

\bibitem{causality}
J.~Pearl, \emph{Causality}.\hskip 1em plus 0.5em minus 0.4em\relax Cambridge university press, 2009.

\bibitem{linliangpami}
Y.~Liu, G.~Li, and L.~Lin, ``Cross-modal causal relational reasoning for event-level visual question answering,'' \emph{TPAMI}, 2023.

\bibitem{granger_event_context}
W.~Zhang, T.~Panum, and e.~a. Jha, Somesh, ``Cause: Learning granger causality from event sequences using attribution methods,'' in \emph{ICML}.\hskip 1em plus 0.5em minus 0.4em\relax PMLR, 2020, pp. 11\,235--11\,245.

\bibitem{granger_event_context2}
D.~W. Gow~Jr and B.~B. Olson, ``Sentential influences on acoustic-phonetic processing: A granger causality analysis of multimodal imaging data,'' \emph{LCN}, vol.~31, no.~7, pp. 841--855, 2016.

\bibitem{xiaoding}
J.~Gao, X.~Ding, B.~Qin, and T.~Liu, ``Is chatgpt a good causal reasoner? a comprehensive evaluation,'' \emph{arXiv preprint arXiv:2305.07375}, 2023.

\bibitem{longtail}
K.~Tang, J.~Huang, and H.~Zhang, ``Long-tailed classification by keeping the good and removing the bad momentum causal effect,'' \emph{NeurIPS}, vol.~33, pp. 1513--1524, 2020.

\bibitem{granger_app}
A.~Tank, I.~Covert, and e.~a. Foti, Nicholas, ``Neural granger causality,'' \emph{TPAMI}, vol.~44, no.~8, pp. 4267--4279, 2021.

\bibitem{videobert}
C.~Sun and e.~a. Myers, Austin, ``Videobert: A joint model for video and language representation learning,'' in \emph{ICCV}, 2019, pp. 7464--7473.

\bibitem{resnet}
K.~He, X.~Zhang, S.~Ren, and J.~Sun, ``Deep residual learning for image recognition,'' in \emph{CVPR}, 2016, pp. 770--778.

\bibitem{caba2015activitynet_ar}
F.~Caba~Heilbron and e.~a. Escorcia, Victor, ``Activitynet: A large-scale video benchmark for human activity understanding,'' in \emph{CVPR}, 2015, pp. 961--970.

\bibitem{overview}
C.~Gong, D.~Yao, and e.~a. Zhang, Chuzhe, ``Causal discovery from temporal data: An overview and new perspectives,'' \emph{arXiv preprint arXiv:2303.10112}, 2023.

\bibitem{clip}
A.~Radford and e.~a. Kim, Jong, ``Learning transferable visual models from natural language supervision,'' in \emph{ICML}, 2021, pp. 8748--8763.

\bibitem{mixtral}
A.~Q. Jiang, A.~Sablayrolles, and e.~a. Roux, Antoine, ``Mixtral of experts,'' \emph{arXiv preprint arXiv:2401.04088}, 2024.

\bibitem{videoLLaVA}
B.~Lin, B.~Zhu, and e.~a. Ye, Yang, ``Video-llava: Learning united visual representation by alignment before projection,'' \emph{arXiv preprint arXiv:2311.10122}, 2023.

\bibitem{mvbench}
K.~Li, Y.~Wang, and e.~a. He, Yinan, ``Mvbench: A comprehensive multi-modal video understanding benchmark,'' in \emph{CVPR}, 2024, pp. 22\,195--22\,206.

\bibitem{longva}
P.~Zhang and e.~a. Zhang, Kaichen, ``Long context transfer from language to vision,'' \emph{arXiv preprint arXiv:2406.16852}, 2024.

\bibitem{qwen25vl}
S.~Bai and e.~a. Chen, Keqin, ``Qwen2. 5-vl technical report,'' \emph{arXiv preprint arXiv:2502.13923}, 2025.

\bibitem{few1}
X.~Wang, W.~Zhu, and e.~a. Saxon, Michael, ``Large language models are latent variable models: Explaining and finding good demonstrations for in-context learning,'' \emph{NeurIPS}, vol.~36, 2024.

\bibitem{few2}
K.~Luo, T.~Zhou, and e.~a. Chen, Yubo, ``Open event causality extraction by the assistance of llm in task annotation, dataset, and method,'' in \emph{COLING}, 2024, pp. 33--44.

\bibitem{few3}
A.~Vashishtha and e.~a. Reddy, Abbavaram~Gowtham, ``Causal inference using llm-guided discovery,'' \emph{arXiv preprint arXiv:2310.15117}, 2023.

\bibitem{deepseekcoder}
Q.~Zhu, D.~Guo, and e.~a. Shao, Zhihong, ``Deepseek-coder-v2: Breaking the barrier of closed-source models in code intelligence,'' \emph{arXiv preprint arXiv:2406.11931}, 2024.

\bibitem{qwen}
J.~Bai, S.~Bai, and e.~a. Chu, Yunfei, ``Qwen technical report,'' \emph{arXiv preprint arXiv:2309.16609}, 2023.

\bibitem{videochat}
D.~Zhu, J.~Chen, and e.~a. Shen, Xiaoqian, ``Minigpt-4: Enhancing vision-language understanding with advanced large language models,'' \emph{arXiv preprint arXiv:2304.10592}, 2023.

\bibitem{minigpt4}
K.~Ataallah, X.~Shen, and e.~a. Abdelrahman, Eslam, ``Minigpt4-video: Advancing multimodal llms for video understanding with interleaved visual-textual tokens,'' \emph{arXiv preprint arXiv:2404.03413}, 2024.

\bibitem{pllava}
L.~Xu, Y.~Zhao, and e.~a. Daquan~Zhou, ``Pllava : Parameter-free llava extension from images to videos for video dense captioning,'' 2024.

\bibitem{siglip}
X.~Zhai, B.~Mustafa, A.~Kolesnikov, and L.~Beyer, ``Sigmoid loss for language image pre-training,'' in \emph{ICCV}, 2023, pp. 11\,975--11\,986.

\bibitem{BLIP2}
J.~Li and e.~a. Li, Dongxu, ``Blip-2: Bootstrapping language-image pre-training with frozen image encoders and large language models,'' in \emph{ICML}, 2023, pp. 19\,730--19\,742.

\bibitem{UniCE}
J.~Gao, C.~Lu, and e.~a. Ding, Xiao, ``Enhancing complex causality extraction via improved subtask interaction and knowledge fusion,'' \emph{arXiv preprint arXiv:2408.03079}, 2024.

\bibitem{SHD1}
V.~Rodr{\'\i}guez-L{\'o}pez and L.~E. Sucar, ``Knowledge transfer for causal discovery,'' \emph{International Journal of Approximate Reasoning}, vol. 143, pp. 1--25, 2022.

\bibitem{SHD2}
K.~Biza, I.~Tsamardinos, and S.~Triantafillou, ``Tuning causal discovery algorithms,'' in \emph{International Conference on Probabilistic Graphical Models}.\hskip 1em plus 0.5em minus 0.4em\relax PMLR, 2020, pp. 17--28.

\bibitem{evaclip}
Q.~Sun, Y.~Fang, and e.~a. Wu, Ledell, ``Eva-clip: Improved training techniques for clip at scale,'' \emph{arXiv preprint arXiv:2303.15389}, 2023.

\bibitem{zhipu}
e.~a. Team, GLM, ``Chatglm: A family of large language models from glm-130b to glm-4 all tools,'' \emph{arXiv e-prints}, pp. arXiv--2406, 2024.

\bibitem{meta2024}
Meta, ``Introducing meta llama 3: The most capable openly available llm to date,'' \url{https://ai.meta.com/blog/meta-llama-3/}, 2024.

\bibitem{univl}
H.~Luo, L.~Ji, and e.~a. Shi, Botian, ``Univl: A unified video and language pre-training model for multimodal understanding and generation,'' \emph{arXiv preprint arXiv:2002.06353}, 2020.

\bibitem{all}
J.~Wang, Y.~Ge, and e.~a. Yan, Rui, ``All in one: Exploring unified video-language pre-training,'' in \emph{CVPR}, 2023, pp. 6598--6608.

\bibitem{languagebind}
B.~Zhu, B.~Lin, and e.~a. Ning, Munan, ``Languagebind: Extending video-language pretraining to n-modality by language-based semantic alignment,'' \emph{arXiv preprint arXiv:2310.01852}, 2023.

\bibitem{activitynetqa}
Z.~Yu, D.~Xu, and e.~a. Yu, Jun, ``Activitynet-qa: A dataset for understanding complex web videos via question answering,'' in \emph{AAAI}, vol.~33, no.~01, 2019, pp. 9127--9134.

\end{thebibliography}

\begin{IEEEbiography}[{\includegraphics[width=1in,height=1.25in,clip,keepaspectratio]{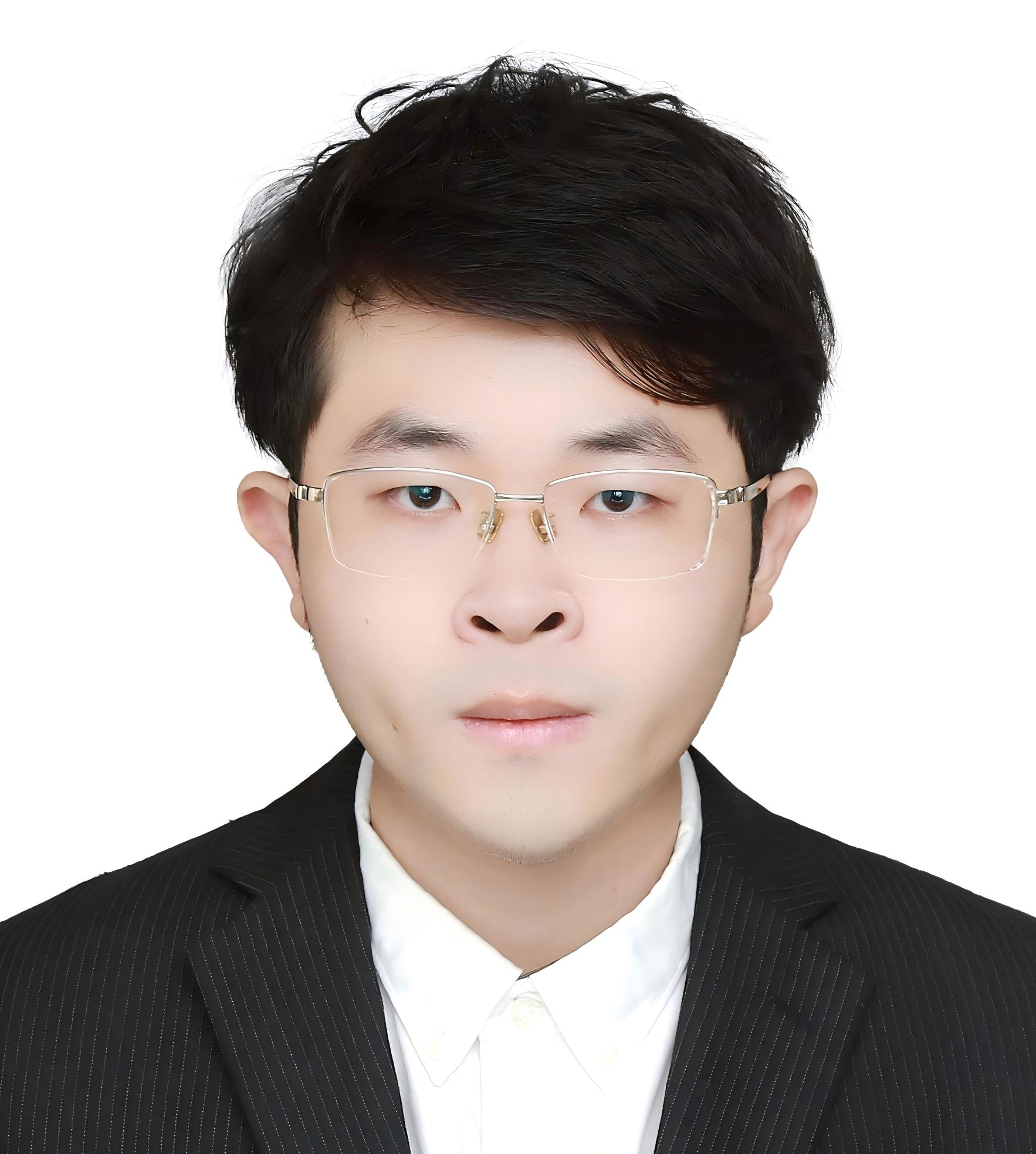}}]{Tieyuan Chen} 
received the Bachelor’s degree in the school of electronic information and electrical engineering from Sichuan University in 2023. He is currently pursuing the Ph.D. degree at Shanghai Jiao Tong University and the Zhongguancun Academy. His research interests include causal discovery, causal reasoning, and video reasoning.
\end{IEEEbiography}

\begin{IEEEbiography}[{\includegraphics[width=1in,height=1.25in,clip,keepaspectratio]{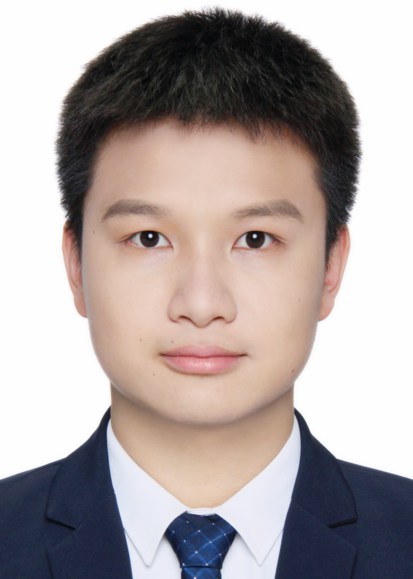}}]{Huabin Liu} 
received the Bachelor’s degree in the school of electronic engineering from Xidian University in 2019. He is currently pursuing the Ph.D. degree at Shanghai Jiao Tong University. His research interests include action understanding, video reasoning, and captioning.
\end{IEEEbiography}

\begin{IEEEbiography}[{\includegraphics[width=1in,height=1.25in,clip,keepaspectratio]{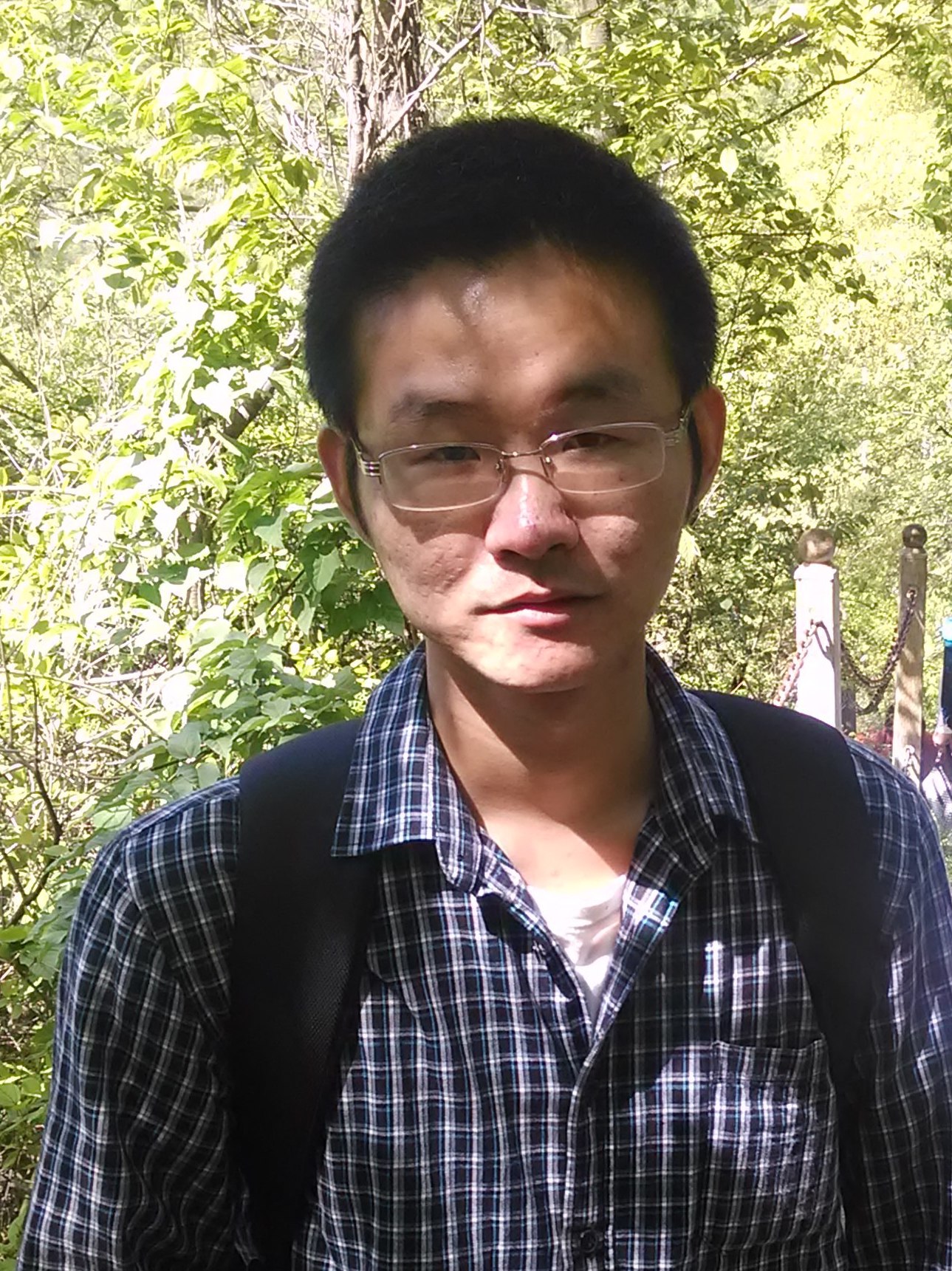}}]{Yi Wang} is a research scientist in Shanghai AI Lab. He received the Ph.D. degree at The Chinese University of Hong Kong in 2021. His current research interests include video understanding and multimodal large language models.
\end{IEEEbiography}

\begin{IEEEbiography}[{\includegraphics[width=1in,height=1.25in,clip,keepaspectratio]{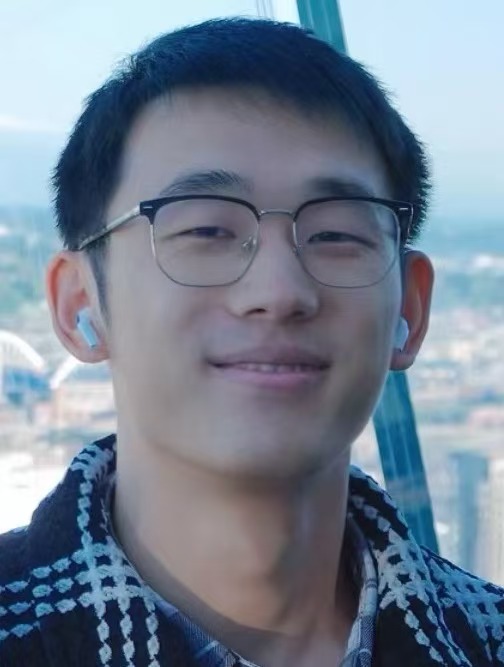}}]{Yihang Chen}
received the Bachelor’s degree from Electronic Information Engineering, Dalian University of Technology, in 2021. He is currently pursuing the Ph.D. degree with the joint program of Shanghai Jiao Tong University and Monash University. His current research interests include 3D vision and data compression techniques.
\end{IEEEbiography}

\begin{IEEEbiography}[{\includegraphics[width=1in,height=1.25in,clip,keepaspectratio]{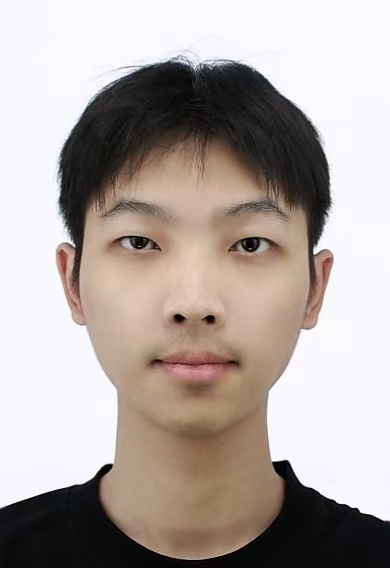}}]{Tianyao He} 
earned his Bachelor's degree from the School of Electronic Information and Electrical Engineering at Shanghai Jiao Tong University and is presently enrolled in a Master's program at the same university. His areas of research focus on video analysis and multimodal comprehension.
\end{IEEEbiography}

\begin{IEEEbiography}[{\includegraphics[width=1in,height=1.3in,clip,keepaspectratio]{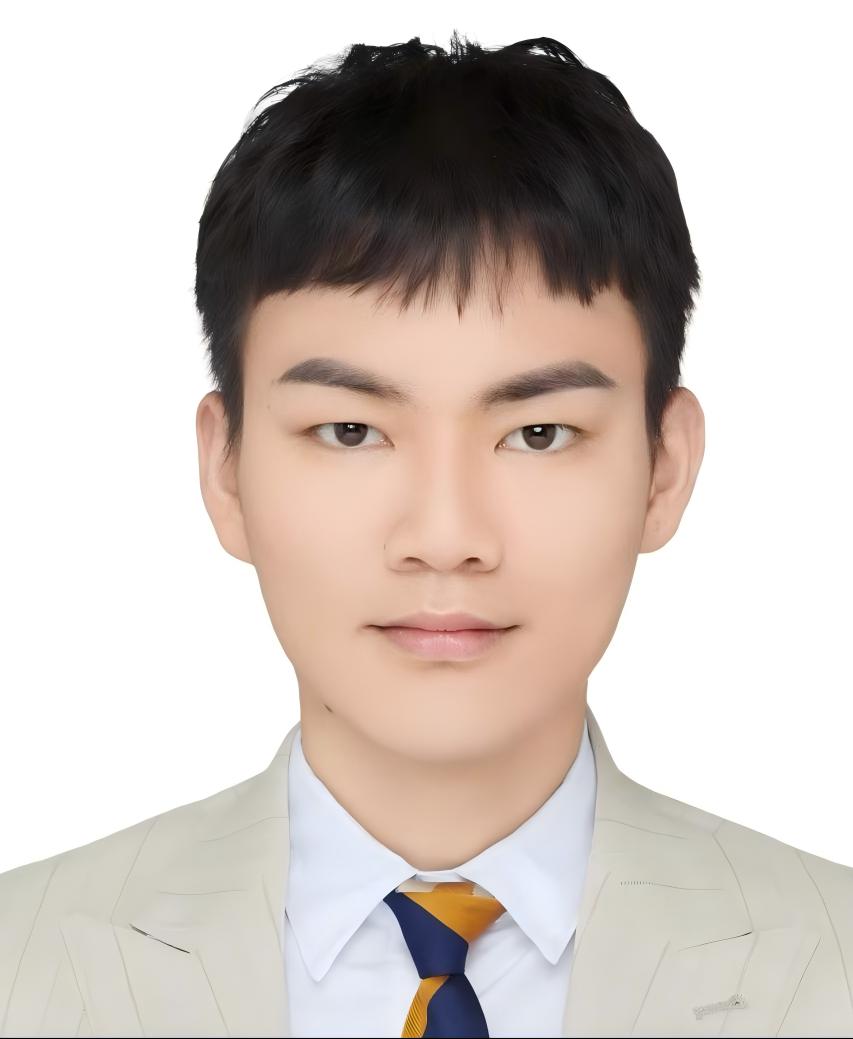}}]{Chaofan Gan}
received the Bachelor's degree in Software Engineering from Huazhong University of Science and Technology, Wuhan, China in 2022. He is currently pursuing the Ph.D. degree in Electronic Information and Electrical Engineering at Shanghai Jiao Tong University, Shanghai, China. His research interests include video understanding, diffusion models, and noise learning.
\end{IEEEbiography}

\begin{IEEEbiography}[{\includegraphics[width=1in,height=1.25in,clip,keepaspectratio]{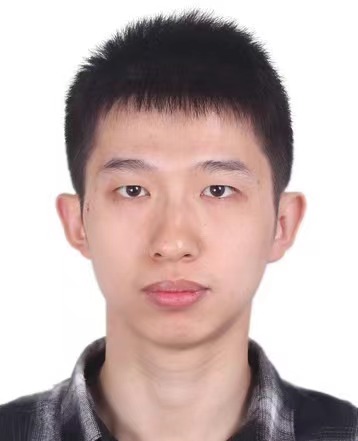}}]{Huanyu He}
received the B.E. and M.E. degrees from Nanjing University in 2016 and Shanghai Jiao Tong University in 2020. He is currently pursuing the Ph.D. degree in the School of Electronic Information and Electrical Engineering at Shanghai Jiao Tong University. His current research interests include machine learning and computer vision, especially on object detection and pedestrian detection.
\end{IEEEbiography}

\begin{IEEEbiography}[{\includegraphics[width=1in,height=1.25in,clip,keepaspectratio]{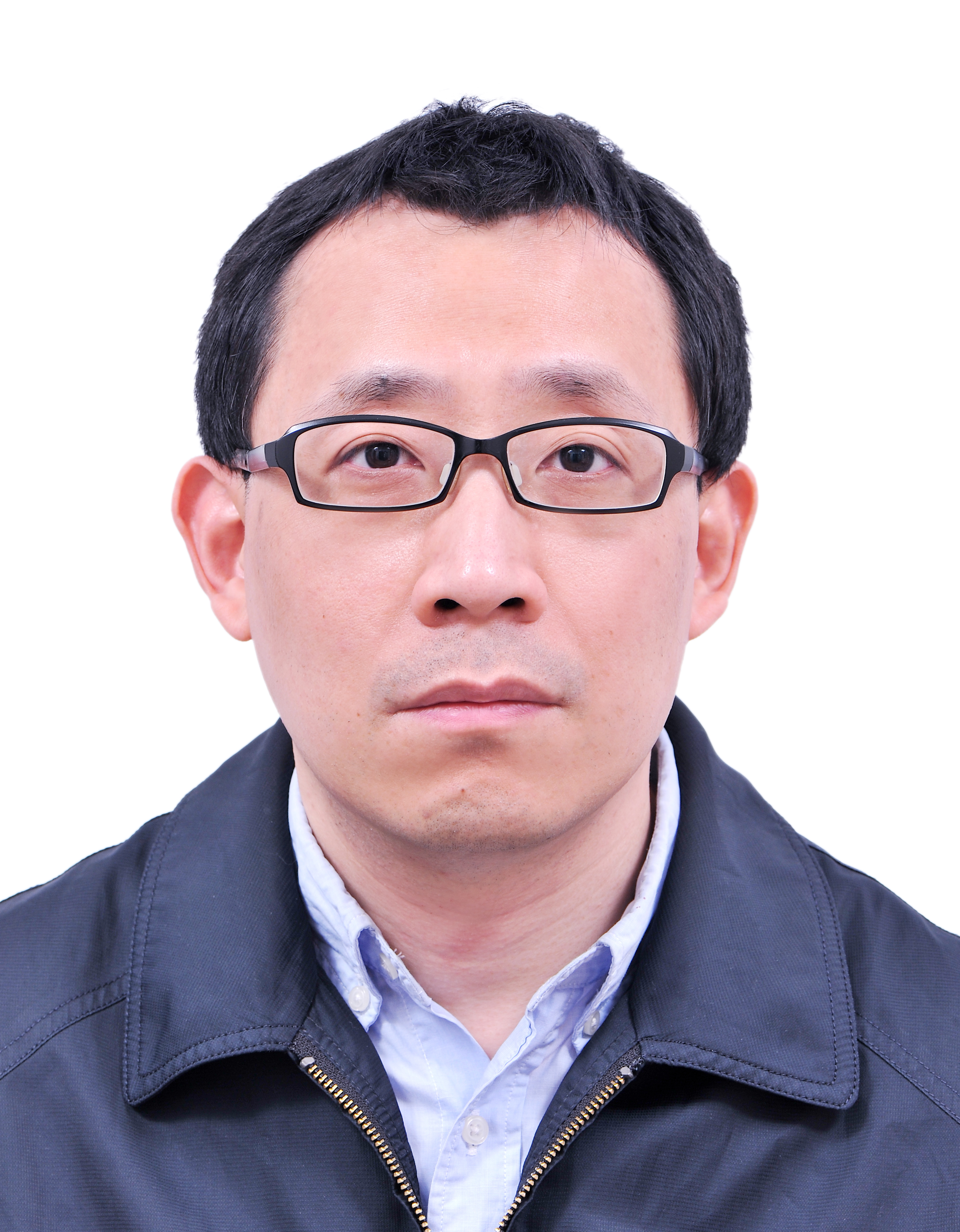}}]{Weiyao Lin}
(Senior Member, IEEE) received the B.E. and M.E. degrees from Shanghai Jiao Tong University, Shanghai, China, in 2003 and 2005, respectively, and the Ph.D. degree from the University of Washington, Seattle, WA, USA, in 2010, all in electrical engineering. He is currently a Professor at the Department of Electronic Engineering, Shanghai Jiao Tong University. He has authored or coauthored more than 100 technical articles on top journals/conferences including the IEEE TRANSACTIONS ON PATTERN ANALYSIS AND MACHINE INTELLIGENCE, the International Journal of Computer Vision, the IEEE TRANSACTIONS ON IMAGE PROCESSING, CVPR, NeurIPS, ICLR, and ICCV. He holds more than 20 patents. His research interests include video/image analysis, computer vision, and video/image processing applications.
\end{IEEEbiography}

\end{document}